\documentclass[11pt]{article}
\usepackage{amsfonts}
\usepackage[preprint]{acl}

\usepackage{times}
\usepackage{latexsym}
\usepackage{amsmath}
\usepackage{cuted}
\usepackage{capt-of} 
\usepackage[T1]{fontenc}

\usepackage[utf8]{inputenc}

\usepackage{microtype}
\usepackage{stfloats}
\usepackage{placeins}    
\usepackage{subcaption} 
\usepackage{float} 
\usepackage{inconsolata}

\usepackage{graphicx}

\title{Layer-wise Positional Bias in Short-Context Language Modeling}

\author{
 \textbf{Maryam Rahimi\textsuperscript{1}} \quad
 \textbf{Mahdi Nouri\textsuperscript{2}} \quad
 \textbf{Yadollah Yaghoobzadeh\textsuperscript{1,2}}
\\
\textsuperscript{1}Tehran Institute for Advanced Studies, Khatam University, Iran\\
\textsuperscript{2}School of Electrical and Computer Engineering, University of Tehran, Iran\\
\texttt{maryamrahimiha@gmail.com}, \textnormal{\texttt{\{mahdi.noori, y.yaghoobzadeh\}@ut.ac.ir}}
 }

\begin{document}
\maketitle

\begin{abstract}
Transformer language models systematically prefer tokens at specific input positions regardless of semantic relevance---a phenomenon known as positional bias. Prior work characterizes this bias in model behavior through performance drops in long-context tasks or in model architecture through attention-based analyses. However, it remains unmeasured how input positions actually drive predictions layer by layer. We introduce a layer conductance framework within a sliding-window design, applied to short-context next-word prediction to isolate model-internal behavior from task and context-window pressure. The resulting layer-wise positional importance profiles are stable across diverse texts and lexical scrambling, confirming they reflect model-internal structure. Characterizing how these profiles evolve across depth, we find recency bias increases monotonically while primacy bias is subtle and diminishes. We also find that this positional bias is not uniform across word types: function words exhibit higher recency bias while content words show higher primacy bias. 
\end{abstract}

\section{Introduction}

Large language models (LLMs) often rely on information from certain input positions more than others, regardless of semantic relevance---a phenomenon known as positional bias. 
Understanding how this bias emerges and evolves across the model's layers is essential for building more reliable language systems.

Prior work has studied this bias from two angles. First, empirical studies show models struggle to use information placed in the middle of long inputs---the ``lost-in-the-middle'' effect, observed across question answering, retrieval, and summarization \cite{shi2023large, Liuetal2024,ko-etal-2020-look}. However, the severity and shape of this effect (e.g., U-shaped profiles versus pure recency) vary substantially across models \cite{zhang-etal-2024-bench, menschikov2025earlytokenbiasmodelspecificlanguagespecific} and context lengths \cite{veseli2025positional} and task demands \cite{Cuconasuetal2025,Salvatoreetal2025}. This variation is expected because the performance degradation measured in these studies is not solely due to the model's intrinsic biases, but is also affected by the difficulty of the task and the pressure of a long context.
Second, theoretically, recent studies show that positional bias is built into the transformer architecture \cite{wu2025emergence, herasimchyk2026residual}. These studies characterize this bias using attention rollout, a propagation-based approximation that tracks how attention weights distribute across positions throughout the layers \cite{abnar-zuidema-2020-quantifying}. However, this approach does not measure the actual contribution of each input position to the model's final prediction \cite{jain2019attention}. What remains uncharacterized is a direct, layer-by-layer measurement of these predictive contributions (see Appendix~\ref{app:related} for extended related work).

We address this gap in short-context next-word prediction---a controlled setting that removes task complexity and context-window pressure, isolating model-internal behavior. We use layer conductance \cite{dhamdhere2018}, a gradient-based attribution method that measures each input's contribution to the final prediction through the hidden states at each layer.
To obtain position-resolved importance scores at each layer, we embed layer conductance within a sliding-window approach that rotates the same set of words through all relative positions. This design also controls for lexical confounds, ensuring that positional patterns are not driven by which words happen to appear at which positions.
Applied to four decoder-only transformers spanning different scales and positional encoding schemes (learned absolute embeddings and RoPE), this yields \emph{layer-wise positional importance profiles}: maps of how each layer distributes its predictive importance across input positions.

We find these profiles are remarkably stable across diverse texts and lexical scrambling ($r > 0.99$), confirming they reflect stable model-internal patterns rather than input content. Tracking how positional importance profiles change across model layers, we show that recency bias increases monotonically while primacy bias is subtle and diminishes.
We further find that this positional bias is not uniform across word types: function words exhibit higher recency bias while content words show higher primacy bias throughout depth. 
Together, these findings provide a concrete empirical characterization of layer-wise positional structure that debiasing methods can directly target.

\section{Methodology}

This section describes how we compute position-resolved layer importance scores and aggregate them to isolate model-internal positional bias.

\begin{figure*}
    \centering
    \includegraphics[width=0.85\linewidth]{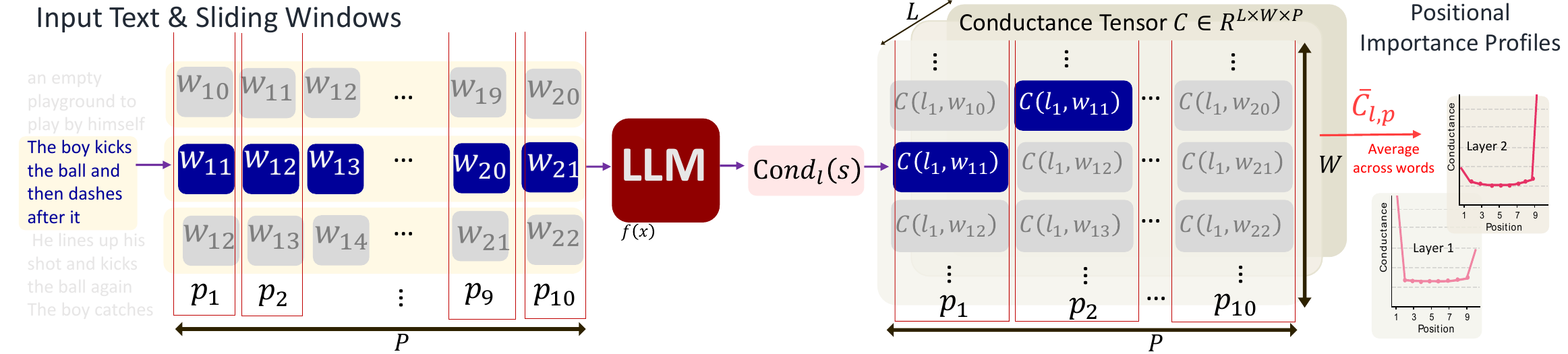}
    \vspace{-3mm}
    \caption{Conductance framework for positional bias. We move sliding windows over input text, extract layer conductance for each word-position pair from each window, store them in tensor $\mathcal{C} \in \mathbb{R}^{L \times W \times P}$, and aggregate across words to obtain positional profiles $\bar{C}_{\ell,p}$.}
    \vspace{-5mm}
    \label{fig:method}
\end{figure*}

\subsection{Attribution via Layer Conductance}

Our primary objective is to quantify how importance is distributed across input positions at each layer. Because raw attention weights 
do not reliably measure contributions to model outputs~\cite{jain2019attention, serrano2019attention}, we adopt 
Layer Conductance~\cite{dhamdhere2018}, a gradient-based attribution method.

Let $h_{\ell,s}$ represent the hidden state vector (hidden dimension $D$) at layer $\ell$ and sequence step $s$. Following \citet{shrikumar2018computationallyefficientmeasuresinternal}, we 
compute layer conductance using the Captum library~\cite{captum_software}, where $f$ denotes the next-token prediction logit:
\vspace{-1mm}
\begin{equation}
\small
C_{\ell}(s) = \sum_{d=1}^{D} \int_{0}^{1} 
\frac{\partial f(h(\alpha))}{\partial h_{\ell,s,d}} \cdot 
\frac{\partial h_{\ell,s,d}(\alpha)}{\partial \alpha} \, d\alpha
\end{equation}

Here $h(\alpha)$ denotes the hidden state at layer $\ell$ when the model receives the interpolated input, varying from a zero-embedding 
baseline to the actual input. The first term measures how sensitive the final prediction is to hidden state slot $s$ at layer $\ell$; 
the second captures how that slot changes along the interpolation path. Each sequence step $s$ corresponds to a single subword token; 
$C_{\ell}(s)$ therefore quantifies the importance of the spatial slot occupied by that token at layer $\ell$ for the final prediction. 

We normalize scores across the sequence to sum to~1, yielding $\widetilde{C}_{\ell}(s)$. Since our analysis operates at the word 
level and surface words may span multiple consecutive subword slots $S(i)$, we obtain word-level scores by summing: 
$C_{\ell}(w_i) = \sum_{s \in S(i)} \widetilde{C}_{\ell}(s)$. 
This aligns importance scores with linguistic units and enables POS-based analysis across models with different tokenizers.

\subsection{Sliding-Window Attribution}

To characterize how each layer distributes predictive importance across relative input positions, we examine word conductance $C_\ell(w)$ across multiple input sequences. Comparing scores across independent sequences, however, would conflate positional with lexical effects, since certain word types systematically appear at specific positions (e.g., sentence-initial tokens). 
To address this confound, we employ a sliding-window approach.
We move a window of length $P = 10$ across each text with a stride of one (Figure~\ref{fig:method}), extracting layer 
conductance for every window. We choose $P = 10$ to enable fine-grained positional analysis while ensuring sufficient context for meaningful next-token prediction; experiments with a longer context of $P = 50$ are provided in Appendix~\ref{app:p50}.

We remove words near the start and end of the text that do not complete the full rotation across all $P$ positions, ensuring each position contains the same set of words. For each word $w$ and layer $\ell$, we collect conductance scores by relative position:

$\small
\mathbf{c}_\ell(w) = 
  (C_{\ell,w,1},\, C_{\ell,w,2},\,\ldots,\, 
   C_{\ell,w,P})
$
\noindent where $C_{\ell,w,p}$ is the word conductance of $w$ at relative position $p$ within
a window.
Collecting these vectors across all words and layers forms tensor $\mathcal{C} \in \mathbb{R}^{L \times W \times P}$ 
($L$ layers, $W$ words, $P$ positions). 

\begin{figure*}[t]
    \centering
    \includegraphics[width=1\linewidth]{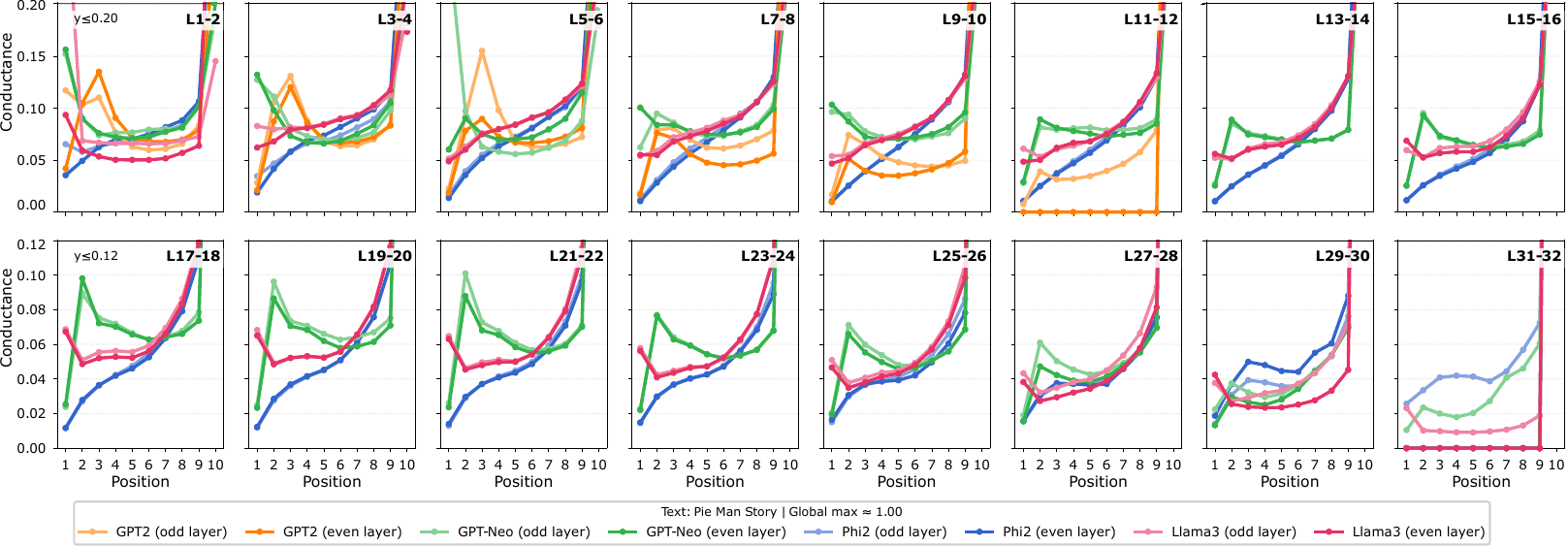}
    \vspace{-5mm}
    \caption{
    Layer-wise positional importance profiles for the \textit{Pie Man} story (Appendix \ref{app:profiles} for other texts). Curves show conductance averaged over words as a function of relative input position; odd layers lighter, even layers darker. Y-axis ranges are adjusted per row (global max = 1.00). All models exhibit a pronounced recency peak growing with depth and a local elevation at initial positions in early layers.
       }
    \vspace{-4mm}
    \label{fig:layer_profiles}
\end{figure*}

\subsection{Positional Importance Profiles}\label{positonal_profile}

For each layer $\ell$, we construct a \textit{positional importance profile}---a vector describing how that layer's predictive importance is distributed across relative input positions---by averaging word conductance across all $W$ words (Figure~\ref{fig:method}, right):
\vspace{-1mm}
\begin{equation}
\small
\bar{C}_{\ell,p} = \frac{1}{W} \sum_{w=1}^{W} 
C_{\ell,w,p}
\label{eq:word_ave}
\end{equation}

\noindent yielding $\small \bar{C}_{\ell} = (\bar{C}_{\ell,1}, \ldots, \bar{C}_{\ell,P})$. 

\noindent\textbf{Consistency} To verify profiles reflect model-internal behavior rather than input content, we compute mean pairwise Pearson correlations between positional importance profiles obtained from the natural texts and the randomized word-shuffled baseline, for every model.

\noindent\textbf{Primacy and recency} To summarize positional structure as scalar measures, we define 
fractions over the first and last 20\% of positions ($\mathcal{P}$ and $\mathcal{R}$):
\vspace{-1mm}
\begin{equation}
\small
\mathrm{PrimFrac}_{\ell} = 
\frac{\sum_{p \in \mathcal{P}} \bar{C}_{\ell,p}}
{\sum_{p=1}^{P} \bar{C}_{\ell,p}}
\qquad
\mathrm{RecFrac}_{\ell} = 
\frac{\sum_{p \in \mathcal{R}} \bar{C}_{\ell,p}}
{\sum_{p=1}^{P} \bar{C}_{\ell,p}}
\label{eq:prim/rec}
\end{equation}
\vspace{-5mm}
\subsection{Word-type Analysis}
To examine whether positional bias differs by word type, we group 
words by type after extracting conductance scores, and compute 
PrimFrac and RecFrac separately for content and function words 
following Equations~\ref{eq:word_ave} and~\ref{eq:prim/rec}. We 
apply FDR-corrected permutation tests to assess whether the 
per-layer difference $\Delta = \text{content} - \text{function}$ 
is significant. POS groupings and statistical details are provided 
in Appendix~\ref{app:wordtype}.
\section{Experimental Setup and Results} \label{sec: results}
\paragraph{Datasets}
To ensure our findings generalize across natural language distributions, we analyze eight texts spanning three distinct genres: narrative (spoken language) \citep{nastase2021narratives}, encyclopedic (factual) \citep{wikipedia_kamali_2025}, and scientific (dense terminology) \citep{thakur2021beir}. We selected these texts to isolate natural language positional effects from the highly rigid, non-linguistic syntax of code or formal mathematics. To decouple positional bias from semantic content, we also include a lexically scrambled control text. Full dataset details are provided in Appendix~\ref{app:datasets}.

\paragraph{Models}\label{sec:models}
We evaluate four autoregressive transformers spanning diverse scales, depths, and, crucially, distinct positional encoding schemes: GPT-2 \cite{radford2019language} and GPT-Neo-2.7B \cite{black2021gptneo} use Absolute Positional Embeddings (APE), whereas Phi-2 \cite{gunasekar2023phi} and Llama-3-8B \cite{meta2024llama3} utilize Rotary Positional Embeddings (RoPE).

\subsection{Positional Profiles and Text Invariance}\label{sec:consistency}

To understand how transformer layers utilize positional information, we aggregate word conductance into layer-wise positional importance profiles (Figure~\ref{fig:layer_profiles}). Before analyzing their specific shapes, we address a critical question: are these patterns driven by the specific input text, or do they reflect inherent model behavior?

By comparing profiles across eight diverse natural texts and a lexically scrambled control, we find that each layer's positional importance profile remains remarkably consistent ($r > 0.99$, Appendix~\ref{app:profiles:consistency}). The persistence of these profiles—even when semantic and syntactic dependencies are destroyed—demonstrates that positional allocation is not driven by linguistic structure, but forms a stable, model-internal pattern.

Across models and layers, profiles reveal three consistent structural zones. 
First, layers maintain a relative local elevation at initial positions; while this diminishes with depth, early tokens retain a baseline level of predictive importance~\cite{xiao2024efficient}.
Second, profiles display a mid-window intersection near position 6, marking a point of cross-model convergence.
Third, all layers exhibit a pronounced peak across recent positions that sharpens with depth. 
These patterns scale to extended contexts: at $P=50$, the mid-window region develops a more pronounced suppression, consistent with the lost-in-the-middle phenomenon (Appendix~\ref{app:p50}).

\subsection{Layer-wise Evolution of Positional Bias}\label{sec:rec-prim_evolution}
\begin{figure}[t]
    \centering
    \includegraphics[width=1\linewidth]{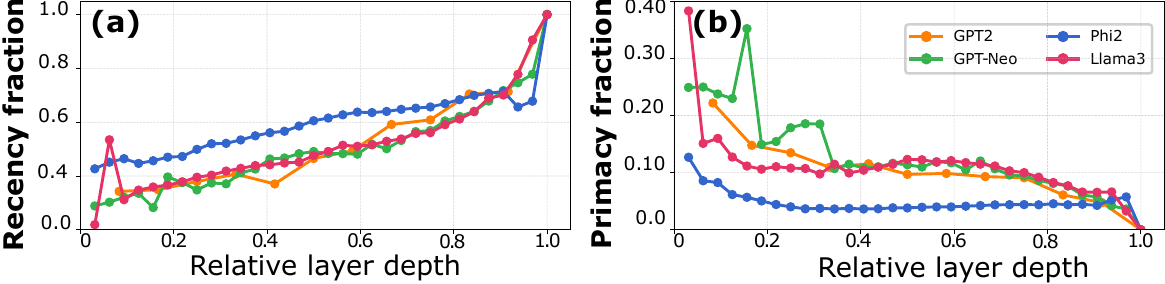}
    \vspace{-6mm}
\caption{Positional bias across layers. (a) Recency fraction increases with depth. (b) Primacy fraction peaks early, then stabilizes. Averaged across all texts.}
    \label{fig:primacy_recency}
    \vspace{-3mm}
\end{figure}

\begin{figure}
    \centering
    \includegraphics[width=1\linewidth]{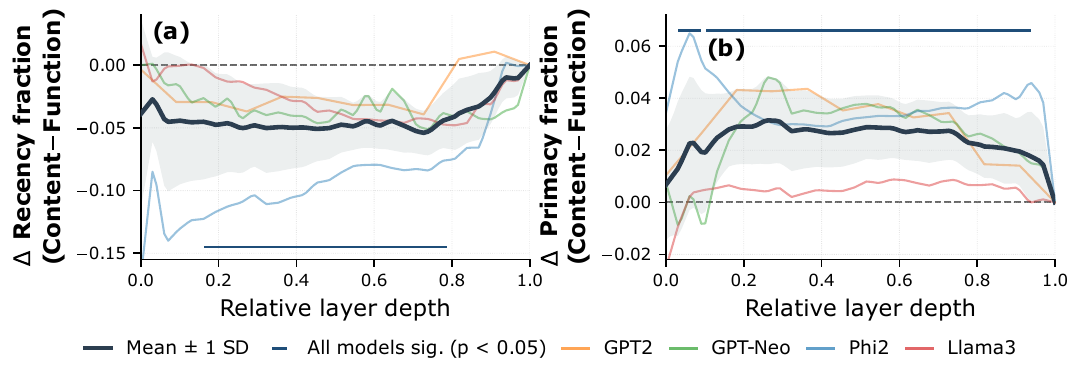}
       \vspace{-6mm}


\caption{Per-layer difference in PrimFrac and RecFrac between content and function words ($\Delta$ = content $-$ function). Words are pooled across all texts (excluding scrambled). Bold line: cross-model mean $\pm$1 SD. Horizontal bar: all four models are significant in the same direction (permutation test, FDR-corrected $p < 0.05$).}

\vspace{-5mm}
     \label{fig:wordtype_delta}
\end{figure}
To quantify this depth-dependent evolution, we track Recency and Primacy fractions across layers (Section~\ref{positonal_profile}, Figure~\ref{fig:primacy_recency}).

Recency bias increases monotonically with depth across all models, from broadly distributed importance in early layers (recency $\approx$ 0.3--0.4) to near-complete concentration on the immediately preceding token in the final layer. In contrast, primacy bias peaks in early layers before stabilizing lower; its magnitude varies by architecture: Llama-3 and GPT-Neo show initial anchoring (primacy $>$ 0.25), Phi-2's primacy is consistently suppressed, and GPT-Neo displays early-layer oscillations, likely reflecting its alternating local/global attention layers~\cite{black2021gptneo}.
To our knowledge, this is the first characterization of how primacy and recency effects evolve across individual model layers. 

\subsection{Positional Bias Across Word Types}

We next ask whether primacy and recency bias differ depending on word type. We compute PrimFrac and RecFrac separately for content words (e.g. nouns, verbs) and function words (e.g. determiners, prepositions), and track the per-layer difference $\Delta = \text{content} - \text{function}$ (Figure~\ref{fig:wordtype_delta}).

Two patterns emerge. First, function words receive higher recency bias than content words ($\Delta\text{Rec} < 0$) across most of the depth range, replicated across all four architectures (FDR-corrected permutation test, $p < 0.05$; Figure~\ref{fig:wordtype_delta}a). Second, content words show higher primacy bias throughout depth ($\Delta\text{Prim} > 0$), consistent across all four models (Figure~\ref{fig:wordtype_delta}b). Both differences converge toward zero at the final layer, as expected: the final layer assigns all importance to the immediately preceding token regardless of word type.

These patterns align with the distinct linguistic roles of the two word classes. Function words serve as local syntactic connectors without independent semantic meaning — their higher recency fractions suggest they matter most to the model when placed near the prediction point. 
Content words carry core semantic meaning—their higher primacy fractions suggest they support contextual integration beyond recent positions, particularly in middle layers which specialize in building higher-level semantic representations~\cite{jawahar2019what, tenney2019bert}.
Together, these results show that positional bias is not uniform and is modulated by word type.

\section{Conclusion}
We characterized layer-wise positional bias in transformer language 
models through a controlled attribution framework, independent of 
task and context-window pressure. The stability of positional 
importance profiles across texts and lexical scrambling confirms that 
these patterns reflect model-internal structure rather than input 
content. Analyzing these profiles reveals recency and primacy biases that are not uniform across 
word types, establishing an empirical baseline for debiasing methods.

\clearpage
\newpage

\section*{Limitations}
This study focuses on positional importance in short-context next-word prediction and is subject to several limitations. 

\paragraph{Short-context scope}
Our short-context setting deliberately removes task and context-window pressure to isolate model-internal behavior. Extending this framework to longer contexts is a natural next step, though computationally demanding: layer conductance requires a full forward-backward pass per window per layer, making large-scale long-context analysis substantially more expensive than standard inference. Comparing profiles obtained here against those under increasing context pressure would directly quantify how these positional patterns are modulated in long-context settings.

\paragraph{Architectural scope}
We study four autoregressive transformers with standard full causal attention. Models with alternative attention mechanisms---such as sliding-window or sparse attention 
layers---require methodological adaptation before layer conductance profiles can be meaningfully interpreted, as these mechanisms alter which positions each layer can attend 
to. Encoder-decoder architectures present a further challenge, since bidirectional attention in the encoder removes the causal constraint central to our analysis. Extending 
the framework to these settings is a promising direction.

\paragraph{Attribution method}
Layer conductance provides principled, gradient-based attribution but remains an approximation of internal computation. The stability of profiles under lexical scrambling ($r > 0.99$) provides evidence that the patterns are robust rather than method-specific; nonetheless, complementary attribution methods may capture aspects of layer behavior not reflected here.

\paragraph{Word-type analysis}
POS-based grouping captures broad word-type tendencies but does not distinguish finer-grained distinctions such as argument structure or lexical semantic classes. Whether depth-dependent sensitivity varies across more fine-grained categories remains open.

\bibliography{custom}

@inproceedings{dhamdhere2018,
  author       = {Kedar Dhamdhere and
                  Mukund Sundararajan and
                  Qiqi Yan},
  title        = {How Important is a Neuron},
  booktitle    = {7th International Conference on Learning Representations, {ICLR} 2019,
                  New Orleans, LA, USA, May 6-9, 2019},
  publisher    = {OpenReview.net},
  year         = {2019},
  url          = {https://openreview.net/forum?id=SylKoo0cKm},
  bibsource    = {dblp computer science bibliography, https://dblp.org}
}

@inproceedings{jain2019attention,
  title={Attention Is Not Explanation},
  author={Jain, Sarthak and Wallace, Byron C.},
  booktitle={Proceedings of NAACL-HLT},
  year={2019}
}

@inproceedings{serrano2019attention,
  title={Is Attention Interpretable?},
  author={Sofia Serrano and Noah A. Smith},
  booktitle={Annual Meeting of the Association for Computational Linguistics},
  year={2019},
  url={https://api.semanticscholar.org/CorpusID:182953113}
}

@inproceedings{gao-etal-2024-insights,
    title = "Insights into {LLM} Long-Context Failures: When Transformers Know but Don{'}t Tell",
    author = "Gao, Muhan  and
      Lu, TaiMing  and
      Yu, Kuai  and
      Byerly, Adam  and
      Khashabi, Daniel",
    editor = "Al-Onaizan, Yaser  and
      Bansal, Mohit  and
      Chen, Yun-Nung",
    booktitle = "Findings of the Association for Computational Linguistics: EMNLP 2024",
    month = nov,
    year = "2024",
    address = "Miami, Florida, USA",
    publisher = "Association for Computational Linguistics",
    url = "https://aclanthology.org/2024.findings-emnlp.447/",
}

@inproceedings{hsieh-etal-2024-found,
    title = "Found in the middle: Calibrating Positional Attention Bias Improves Long Context Utilization",
    author = "Hsieh, Cheng-Yu  and
      Chuang, Yung-Sung  and
      Li, Chun-Liang  and
      Wang, Zifeng  and
      Le, Long  and
      Kumar, Abhishek  and
      Glass, James  and
      Ratner, Alexander  and
      Lee, Chen-Yu  and
      Krishna, Ranjay  and
      Pfister, Tomas",
    editor = "Ku, Lun-Wei  and
      Martins, Andre  and
      Srikumar, Vivek",
    booktitle = "Findings of the Association for Computational Linguistics: ACL 2024",
    month = aug,
    year = "2024",
    address = "Bangkok, Thailand",
    publisher = "Association for Computational Linguistics",
    url = "https://aclanthology.org/2024.findings-acl.890/",
}

@inproceedings{ko-etal-2020-look,
    title = "Look at the First Sentence: Position Bias in Question Answering",
    author = "Ko, Miyoung  and
      Lee, Jinhyuk  and
      Kim, Hyunjae  and
      Kim, Gangwoo  and
      Kang, Jaewoo",
    editor = "Webber, Bonnie  and
      Cohn, Trevor  and
      He, Yulan  and
      Liu, Yang",
    booktitle = "Proceedings of the 2020 Conference on Empirical Methods in Natural Language Processing (EMNLP)",
    month = nov,
    year = "2020",
    address = "Online",
    publisher = "Association for Computational Linguistics",
    url = "https://aclanthology.org/2020.emnlp-main.84/",
}

@inproceedings{pascual-etal-2021-telling,
    title = "Telling {BERT}{'}s Full Story: from Local Attention to Global Aggregation",
    author = "Pascual, Damian  and
      Brunner, Gino  and
      Wattenhofer, Roger",
    editor = "Merlo, Paola  and
      Tiedemann, Jorg  and
      Tsarfaty, Reut",
    booktitle = "Proceedings of the 16th Conference of the European Chapter of the Association for Computational Linguistics: Main Volume",
    month = apr,
    year = "2021",
    address = "Online",
    publisher = "Association for Computational Linguistics",
    url = "https://aclanthology.org/2021.eacl-main.9/"
}

@misc{hou2023decodinglayersaliencylanguage,
      title={Decoding Layer Saliency in Language Transformers}, 
      author={Elizabeth M. Hou and Gregory Castanon},
      year={2023},
      eprint={2308.05219},
      archivePrefix={arXiv},
      primaryClass={cs.CL},
      url={https://arxiv.org/abs/2308.05219}, 
}

@misc{ikeda2025layerwiseimportanceanalysisfeedforward,
      title={Layerwise Importance Analysis of Feed-Forward Networks in Transformer-based Language Models}, 
      author={Wataru Ikeda and Kazuki Yano and Ryosuke Takahashi and Jaesung Lee and Keigo Shibata and Jun Suzuki},
      year={2025},
      eprint={2508.17734},
      archivePrefix={arXiv},
      primaryClass={cs.CL},
      url={https://arxiv.org/abs/2508.17734}, 
}

@inproceedings{radford2019language,
  title={Language Models are Unsupervised Multitask Learners},
  author={Alec Radford and Jeff Wu and Rewon Child and David Luan and Dario Amodei and Ilya Sutskever},
  year={2019},
  url={https://api.semanticscholar.org/CorpusID:160025533}
}

@inproceedings{black2021gptneo,
  title={GPT-Neo: Large Scale Autoregressive Language Modeling with Mesh-Tensorflow},
  author={Sid Black and Leo Gao and Phil Wang and Connor Leahy and Stella Biderman},
  year={2021},
  url={https://api.semanticscholar.org/CorpusID:245758737}
}

@article{gunasekar2023phi,
  title={Textbooks Are All You Need},
  author={Suriya Gunasekar and Yi Zhang and Jyoti Aneja and Caio C'esar Teodoro Mendes and Allison Del Giorno and Sivakanth Gopi and Mojan Javaheripi and Piero Kauffmann and Gustavo de Rosa and Olli Saarikivi and Adil Salim and S. Shah and Harkirat Singh Behl and Xin Wang and S{\'e}bastien Bubeck and Ronen Eldan and Adam Tauman Kalai and Yin Tat Lee and Yuan-Fang Li},
  journal={ArXiv},
  year={2023},
  volume={abs/2306.11644},
  url={https://api.semanticscholar.org/CorpusID:259203998}
}

@inproceedings{shi2023large,
  title={Large Language Models Can Be Easily Distracted by Irrelevant Context},
  author={Freda Shi and Xinyun Chen and Kanishka Misra and Nathan Scales and David Dohan and Ed H. Chi and Nathanael Scharli and Denny Zhou},
  booktitle={International Conference on Machine Learning},
  year={2023},
  url={https://api.semanticscholar.org/CorpusID:256459776}
}

@article{xiao2024efficient,
  title={Efficient Streaming Language Models with Attention Sinks},
  author={Guangxuan Xiao and Yuandong Tian and Beidi Chen and Song Han and Mike Lewis},
  journal={ArXiv},
  year={2023},
  volume={abs/2309.17453},
  url={https://api.semanticscholar.org/CorpusID:263310483}
}

@inproceedings{han2024lm,
  title={LM-Infinite: Zero-Shot Extreme Length Generalization for Large Language Models},
  author={Chi Han and Qifan Wang and Hao Peng and Wenhan Xiong and Yu Chen and Heng Ji and Sinong Wang},
  booktitle={North American Chapter of the Association for Computational Linguistics},
  year={2023},
  url={https://api.semanticscholar.org/CorpusID:268357635}
}

@inproceedings{meta2024llama3,
  title={The Llama 3 Herd of Models},
  author={Abhimanyu Dubey and Abhinav Jauhri and Abhinav Pandey and Abhishek Kadian and Ahmad Al-Dahle and Aiesha Letman and Akhil Mathur and Alan Schelten and Amy Yang and Angela Fan and Anirudh Goyal and Anthony S. Hartshorn and Aobo Yang and Archi Mitra and Archie Sravankumar and Artem Korenev and Arthur Hinsvark and Arun Rao and Aston Zhang and Aur'elien Rodriguez and Austen Gregerson and Ava Spataru and Baptiste Rozi{\`e}re and Bethany M. Biron and Binh Tang and Bobbie Chern and Char-lotte Caucheteux and Chaya Nayak and Chloe Bi and Chris Marra and Chris McConnell and Christian Keller and Christophe Touret and Chunyang Wu and Corinne Wong and Cristian Canton Ferrer and Cyrus Nikolaidis and Damien Allonsius and Daniel J. Song and Danielle Pintz and Danny Livshits and David Esiobu and Dhruv Choudhary and Dhruv Mahajan and Diego Garcia-Olano and Diego Perino and Dieuwke Hupkes and Egor Lakomkin and Ehab A. AlBadawy and E I Lobanova and Emily Dinan and Eric Michael Smith and Filip Radenovic and Frank Zhang and Gabriel Synnaeve and Gabrielle Lee and Georgia Lewis Anderson and Graeme Nail and Gr{\'e}goire Mialon and Guanglong Pang and Guillem Cu-curell and Hailey Nguyen and Hannah Korevaar and Hu Xu and Hugo Touvron and Iliyan Zarov and Imanol Arrieta Ibarra and Isabel M. Kloumann and Ishan Misra and Ivan Evtimov and Jade Copet and Jaewon Lee and Jan Geffert and Jana Vranes and Jason Park and Jay Mahadeokar and Jeet Shah and Jelmer van der Linde and Jennifer Billock and Jenny Hong and Jenya Lee and Jeremy Fu and Jianfeng Chi and Jianyu Huang and Jiawen Liu and Jie Wang and Jiecao Yu and Joanna Bitton and Joe Spisak and Jongsoo Park and Joseph Rocca and Joshua Johnstun and Joshua Saxe and Ju-Qing Jia and Kalyan Vasuden Alwala and K. Upasani and Kate Plawiak and Keqian Li and Kenneth Heafield and Kevin R. Stone and Khalid El-Arini and Krithika Iyer and Kshitiz Malik and Kuen-ley Chiu and Kunal Bhalla and Lauren Rantala-Yeary and Laurens van der Maaten and Lawrence Chen and Liang Tan and Liz Jenkins and Louis Martin and Lovish Madaan and Lubo Malo and Lukas Blecher and Lukas Landzaat and Luke de Oliveira and Madeline Muzzi and Ma-hesh Pasupuleti and Mannat Singh and Manohar Paluri and Marcin Kardas and Mathew Oldham and Mathieu Rita and Maya Pavlova and Melissa Hall Melanie Kambadur and Mike Lewis and Min Si and Mitesh Kumar Singh and Mona Hassan and Naman Goyal and Narjes Torabi and Nikolay Bash-lykov and Nikolay Bogoychev and Niladri S. Chatterji and Olivier Duchenne and Onur cCelebi and Patrick Alrassy and Pengchuan Zhang and Pengwei Li and Petar Vasi{\'c} and Peter Weng and Prajjwal Bhargava and Pratik Dubal and Praveen Krishnan and Punit Singh Koura and Puxin Xu and Qing He and Qingxiao Dong and R. S. Mughil Srinivasan and Raj Ganapathy and Ramon Calderer and Ricardo Silveira Cabral and Robert Stojnic and Roberta Raileanu and Rohit Girdhar and Rohit Patel and Romain Sauvestre and Ron-nie Polidoro and Roshan Sumbaly and Ross Taylor and Ruan Silva and Rui Hou and Rui Wang and Saghar Hosseini and Sa-hana Chennabasappa and Sanjay Singh and Sean Bell and Seohyun Sonia Kim and Sergey Edunov and Shaoliang Nie and Sharan Narang and Sharath Chandra Raparthy and Sheng Shen and Shengye Wan and Shruti Bhosale and Shun Zhang and Simon Vandenhende and Soumya Batra and Spencer Whit-man and Sten Sootla and St{\'e}phane Collot and Suchin Gururangan and Sydney Borodinsky and Tamar Herman and Tara Fowler and Tarek Sheasha and Thomas Georgiou and Thomas Scialom and Tobias Speckbacher and Todor Mihaylov and Tong Xiao and Ujjwal Karn and Vedanuj Goswami and Vibhor Gupta and Vignesh Ramanathan and Viktor Kerkez and Vincent Gonguet and Virginie Do and Vish Vogeti and Vladan Petrovic and Weiwei Chu and Wenhan Xiong and Wenyin Fu and Whit-ney Meers and Xavier Martinet and Xiaodong Wang and Xiaoqing Ellen Tan and Xinfeng Xie and Xuchao Jia and Xuewei Wang and Yaelle Goldschlag and Yashesh Gaur and Yasmine Babaei and Yiqian Wen and Yiwen Song and Yuchen Zhang and Yue Li and Yuning Mao and Zacharie Delpierre Coudert and Zhengxu Yan and Zhengxing Chen and Zoe Papakipos and Aaditya K. Singh and Aaron Grattafiori and Abha Jain and Adam Kelsey and Adam Shajnfeld and Adi Gangidi and Adolfo Victoria and Ahuva Goldstand and Ajay Menon and Ajay Sharma and Alex Boesenberg and Alex Vaughan and Alexei Baevski and Allie Fein-stein and Amanda Kallet and Amit Sangani and Anam Yunus and Andrei Lupu and Andres Alvarado and Andrew Caples and Andrew Gu and Andrew Ho and Andrew Poulton and Andrew Ryan and Ankit Ramchandani and Annie Franco and Aparajita Saraf and Arkabandhu Chowdhury and Ashley Gabriel and Ashwin R. Bharambe and Assaf Eisenman and Azadeh Yazdan and Beau James and Ben Maurer and Benjamin Leonhardi and Po-Yao (Bernie) Huang and Beth Loyd and Beto de Paola and Bhargavi Paranjape and Bing Liu and Bo Wu and Boyu Ni and Braden Hancock and Bram Wasti and Brandon Spence and Brani Stojkovic and Brian Gamido and Britt Montalvo and Carl Parker and Carly Burton and Catalina Mejia and Changhan Wang and Changkyu Kim and Chao Zhou and Chester Hu and Ching-Hsiang Chu and Chris Cai and Chris Tindal and Christoph Feichtenhofer and Damon Civin and Dana Beaty and Daniel Kreymer and Shang-Wen Li and Danny Wyatt and David Adkins and David Xu and Davide Testuggine and Delia David and Devi Parikh and Diana Liskovich and Didem Foss and Dingkang Wang and Duc Le and Dustin Holland and Edward Dowling and Eissa Jamil and Elaine Montgomery and Eleonora Presani and Emily Hahn and Emily Wood and Erik Brinkman and Esteban Arcaute and Evan Dunbar and Evan Smoth-ers and Fei Sun and Felix Kreuk and Feng Tian and Firat Ozgenel and Francesco Caggioni and Francisco (Paco) Guzm{\'a}n and Frank J. Kanayet and Frank Seide and Gabriela Medina Florez and Gabriella Schwarz and Gada Badeer and Georgia Swee and Gil Halpern and Govind Thattai and Grant Herman and Grigory Sizov and Guangyi Zhang and Guna Lakshminarayanan and Hamid Shojanazeri and Han Zou and Hannah Wang and Han Zha and Haroun Habeeb and Harrison Rudolph and Helen Suk and Henry Aspegren and Hunter Goldman and Igor Molybog and Igor Tufanov and Irina-Elena Veliche and Itai Gat and Jake Weissman and James Geboski and James Kohli and Japhet Asher and Jean-Baptiste Gaya and Jeff Marcus and Jeff Tang and Jennifer Chan and Jenny Zhen and Jeremy Reizenstein and Jeremy Teboul and Jessica Zhong and Jian Jin and Jingyi Yang and Joe Cummings and Jon Carvill and Jon Shepard and Jonathan McPhie and Jonathan Torres and Josh Ginsburg and Junjie Wang and Kaixing(Kai) Wu and U KamHou and Karan Saxena and Karthik Prasad and Kartikay Khandelwal and Katay-oun Zand and Kathy Matosich and Kaushik Veeraraghavan and Kelly Michelena and Keqian Li and Kun Huang and Kunal Chawla and Kushal Lakhotia and Kyle Huang and Lailin Chen and Lakshya Garg and A Lavender and Leandro Silva and Lee Bell and Lei Zhang and Liangpeng Guo and Licheng Yu and Liron Moshkovich and Luca Wehrstedt and Madian Khabsa and Manav Avalani and Manish Bhatt and Maria Tsimpoukelli and Martynas Mankus and Matan Hasson and Matthias Lennie and Matthias Reso and Maxim Groshev and Maxim Naumov and Maya Lathi and Meghan Keneally and Michael L. Seltzer and Michal Valko and Michelle Re-strepo and Mihir Patel and Mik Vyatskov and Mikayel Samvelyan and Mike Clark and Mike Macey and Mike Hang Wang and Miquel Jubert Hermoso and Mo Metanat and Mohammad Rastegari and Mun-ish Bansal and Nandhini Santhanam and Natascha Parks and Natasha White and Navy-ata Bawa and Nayan Singhal and Nick Egebo and Nicolas Usunier and Nikolay Pavlovich Laptev and Ning Dong and Ning Zhang and Norman Cheng and Oleg Chernoguz and Olivia Hart and Omkar Salpekar and Ozlem Kalinli and Parkin Kent and Parth Parekh and Paul Saab and Pavan Balaji and Pe-dro Rittner and Philip Bontrager and Pierre Roux and Piotr Doll{\'a}r and Polina Zvyagina and Prashant Ratanchandani and Pritish Yuvraj and Qian Liang and Rachad Alao and Rachel Rodriguez and Rafi Ayub and Raghotham Murthy and Raghu Nayani and Rahul Mitra and Raymond Li and Rebekkah Hogan and Robin Battey and Rocky Wang and Ro-han Maheswari and Russ Howes and Ruty Rinott and Sai Jayesh Bondu and Samyak Datta and Sara Chugh and Sara Hunt and Sargun Dhillon and S. Yu. Sidorov and Satadru Pan and Saurabh Verma and Seiji Yamamoto and Sharadh Ramaswamy and Shaun Lindsay and Sheng Feng and Shenghao Lin and Shengxin Zha and Shiva Shankar and Shuqiang Zhang and Sinong Wang and Sneha Agarwal and Soji Sajuyigbe and Soumith Chintala and Stephanie Max and Stephen Chen and Steve Kehoe and Steve Satterfield and Sudarshan Govindaprasad and Sumit Kumar Gupta and Sung-Bae Cho and Sunny Virk and Suraj Subramanian and Sy Choudhury and Sydney Goldman and Tal Remez and Tamar Glaser and Tamara Best and Thilo Kohler and Thomas Robinson and Tianhe Li and Tianjun Zhang and Tim Matthews and Timothy Chou and Tzook Shaked and Varun Vontimitta and Victoria O Ajayi and Victoria Montanez and Vijai Mohan and Vinay Kumar and Vishal Mangla and Vlad Ionescu and Vlad Andrei Poenaru and Vlad T. Mihailescu and Vladimir Ivanov and Wei Li and Wenchen Wang and Wenwen Jiang and Wes Bouaziz and Will Constable and Xia Tang and Xiaofang Wang and Xiaojian Wu and Xiaolan Wang and Xide Xia and Xilun Wu and Xinbo Gao and Yanjun Chen and Ye Hu and Ye Jia and Ye Qi and Yenda Li and Yilin Zhang and Ying Zhang and Yossi Adi and Youngjin Nam and Yu Wang and Yuchen Hao and Yundi Qian and Yuzi He and Zach Rait and Zachary DeVito and Zef Rosnbrick and Zhaoduo Wen and Zhenyu Yang and Zhiwei Zhao},
  year={2024},
  url={https://api.semanticscholar.org/CorpusID:271571434}
}

@inproceedings{
wang2025eliminating,
title={Eliminating Position Bias of Language Models: A Mechanistic Approach},
author={Ziqi Wang and Hanlin Zhang and Xiner Li and Kuan-Hao Huang and Chi Han and Shuiwang Ji and Sham M. Kakade and Hao Peng and Heng Ji},
booktitle={The Thirteenth International Conference on Learning Representations},
year={2025},
url={https://openreview.net/forum?id=fvkElsJOsN}
}

@misc{marks2025sparsefeaturecircuitsdiscovering,
      title={Sparse Feature Circuits: Discovering and Editing Interpretable Causal Graphs in Language Models}, 
      author={Samuel Marks and Can Rager and Eric J. Michaud and Yonatan Belinkov and David Bau and Aaron Mueller},
      year={2025},
      eprint={2403.19647},
      archivePrefix={arXiv},
      primaryClass={cs.LG},
      url={https://arxiv.org/abs/2403.19647}, 
}

@inproceedings{ferrando-etal-2022-measuring,
    title = "Measuring the Mixing of Contextual Information in the Transformer",
    author = "Ferrando, Javier  and
      G{\'a}llego, Gerard I.  and
      Costa-juss{\`a}, Marta R.",
    editor = "Goldberg, Yoav  and
      Kozareva, Zornitsa  and
      Zhang, Yue",
    booktitle = "Proceedings of the 2022 Conference on Empirical Methods in Natural Language Processing",
    month = dec,
    year = "2022",
    address = "Abu Dhabi, United Arab Emirates",
    publisher = "Association for Computational Linguistics",
    url = "https://aclanthology.org/2022.emnlp-main.595/"
}

@article{wu2025emergence,
  title={On the Emergence of Position Bias in Transformers},
  author={Xinyi Wu and Yifei Wang and Stefanie Jegelka and Ali Jadbabaie},
  journal={ArXiv},
  year={2025},
  volume={abs/2502.01951},
  url={https://api.semanticscholar.org/CorpusID:276107602}
}

@inproceedings{ravaut-etal-2024-context,
    title = "On Context Utilization in Summarization with Large Language Models",
    author = "Ravaut, Mathieu  and
      Sun, Aixin  and
      Chen, Nancy  and
      Joty, Shafiq",
    editor = "Ku, Lun-Wei  and
      Martins, Andre  and
      Srikumar, Vivek",
    booktitle = "Proceedings of the 62nd Annual Meeting of the Association for Computational Linguistics (Volume 1: Long Papers)",
    month = aug,
    year = "2024",
    address = "Bangkok, Thailand",
    publisher = "Association for Computational Linguistics",
    url = "https://aclanthology.org/2024.acl-long.153/"
}

@misc{menschikov2025earlytokenbiasmodelspecificlanguagespecific,
      title={Beyond Early-Token Bias: Model-Specific and Language-Specific Position Effects in Multilingual LLMs}, 
      author={Mikhail Menschikov and Alexander Kharitonov and Maiia Kotyga and Vadim Porvatov and Anna Zhukovskaya and David Kagramanyan and Egor Shvetsov and Evgeny Burnaev},
      year={2025},
      eprint={2505.16134},
      archivePrefix={arXiv},
      primaryClass={cs.CL},
      url={https://arxiv.org/abs/2505.16134}, 
}

@inproceedings{veseli2025positional,
  title={Positional Biases Shift as Inputs Approach Context Window Limits},
  author={Veseli, Blerta and Chibane, Julian and Toneva, Mariya and Koller, Alexander},
  booktitle={First Conference on Language Modeling},
  year={2025},
  url={https://openreview.net/forum?id=YourPaperID} 
}

@misc{kongmanee2025unravelingtokenpredictionrefinement,
      title={Unraveling Token Prediction Refinement and Identifying Essential Layers in Language Models}, 
      author={Jaturong Kongmanee},
      year={2025},
      eprint={2501.15054},
      archivePrefix={arXiv},
      primaryClass={cs.CL},
      url={https://arxiv.org/abs/2501.15054}, 
}

@inproceedings{
thakur2021beir,
title={{BEIR}: A Heterogeneous Benchmark for Zero-shot Evaluation of Information Retrieval Models},
author={Nandan Thakur and Nils Reimers and Andreas R{\"u}ckl{\'e} and Abhishek Srivastava and Iryna Gurevych},
booktitle={Thirty-fifth Conference on Neural Information Processing Systems Datasets and Benchmarks Track (Round 2)},
year={2021},
url={https://openreview.net/forum?id=wCu6T5xFjeJ}
}

@misc{wikipedia_kamali_2025,
    author       = { Omar Kamali and Omneity Labs },
    title        = { Wikipedia Monthly },
    year         = { 2025 },
    url          = { https://huggingface.co/datasets/omarkamali/wikipedia-monthly },
    doi          = { 10.57967/hf/6575 },
    publisher    = { Hugging Face }
}

@software{captum_software,
  author = {Kokhlikyan, Narine and Miglani, Vivek and Martin, Miguel and Wang, Edward},
  title = {Captum: A unified and generic model interpretability library for PyTorch},
 doi = {https://doi.org/10.48550/ARXIV.2009.07896},
  year = {2023},
  publisher = {GitHub},
  journal = {GitHub repository},
  howpublished = {\url{https://github.com/pytorch/captum}}
}

@article{nastase2021narratives,
  title={The “Narratives” fMRI dataset for evaluating models of naturalistic language comprehension},
  author={Samuel A. Nastase and Yun-Fei Liu and Hanna Hillman and Asieh Zadbood and Liat Hasenfratz and Neggin Keshavarzian and Janice Chen and Christopher John Honey and Yaara Yeshurun and Mor Regev and Mai Nguyen and Claire H. C. Chang and Christopher A. Baldassano and Olga Lositsky and Erez Simony and Michael A. Chow and Yuan Chang Leong and Paula P. Brooks and Emily T. Micciche and Gina Choe and Ariel Goldstein and Tamara Vanderwal and Yaroslav O. Halchenko and Kenneth A. Norman and Uri Hasson},
  journal={Scientific Data},
  year={2021},
  volume={8},
  url={https://api.semanticscholar.org/CorpusID:238218053}
}

@misc{shrikumar2018computationallyefficientmeasuresinternal,
      title={Computationally Efficient Measures of Internal Neuron Importance}, 
      author={Avanti Shrikumar and Jocelin Su and Anshul Kundaje},
      year={2018},
      eprint={1807.09946},
      archivePrefix={arXiv},
      primaryClass={cs.LG},
      url={https://arxiv.org/abs/1807.09946}, 
}

@misc{herasimchyk2026residual,
      title={A Residual-Aware Theory of Position Bias in Transformers}, 
      author={Hanna Herasimchyk and Robin Labryga and Tomislav Prusina and Sören Laue},
      year={2026},
      eprint={2602.16837},
      archivePrefix={arXiv},
      primaryClass={cs.LG},
      url={https://arxiv.org/abs/2602.16837}, 
}

@inproceedings{zhang-etal-2024-bench,
    title = "$\infty${B}ench: Extending Long Context Evaluation Beyond 100{K} Tokens",
    author = "Zhang, Xinrong  and
      Chen, Yingfa  and
      Hu, Shengding  and
      Xu, Zihang  and
      Chen, Junhao  and
      Hao, Moo  and
      Han, Xu  and
      Thai, Zhen  and
      Wang, Shuo  and
      Liu, Zhiyuan  and
      Sun, Maosong",
    editor = "Ku, Lun-Wei  and
      Martins, Andre  and
      Srikumar, Vivek",
    booktitle = "Proceedings of the 62nd Annual Meeting of the Association for Computational Linguistics (Volume 1: Long Papers)",
    month = aug,
    year = "2024",
    address = "Bangkok, Thailand",
    publisher = "Association for Computational Linguistics",
    url = "https://aclanthology.org/2024.acl-long.814/",
    doi = "10.18653/v1/2024.acl-long.814",
}

@inproceedings{jawahar2019what,
    title = "What Does {BERT} Learn about the Structure of Language?",
    author = "Jawahar, Ganesh and Sagot, Beno{\^\i}t and Seddah, Djam{\'e}",
    booktitle = "Proceedings of the 57th Annual Meeting of the Association for Computational Linguistics",
    month = jul,
    year = "2019",
    address = "Florence, Italy",
    publisher = "Association for Computational Linguistics",
    url = "https://aclanthology.org/P19-1356",
    doi = "10.18653/v1/P19-1356",
    pages = "3651--3657"
}

@inproceedings{tenney2019bert,
    title = "{BERT} Rediscovers the Classical {NLP} Pipeline",
    author = "Tenney, Ian and Das, Dipanjan and Pavlick, Ellie",
    booktitle = "Proceedings of the 57th Annual Meeting of the Association for Computational Linguistics",
    month = jul,
    year = "2019",
    address = "Florence, Italy",
    publisher = "Association for Computational Linguistics",
    url = "https://aclanthology.org/P19-1452",
    doi = "10.18653/v1/P19-1452",
    pages = "4593--4601"
}

@article{Liuetal2024,
  author  = {Liu, Nelson F. and Lin, Kevin and Hewitt, John and Paranjape, Ashwin and Bevilacqua, Michele and Petroni, Fabio and Liang, Percy},
  title   = {Lost in the Middle: How Language Models Use Long Contexts},
  journal = {Transactions of the Association for Computational Linguistics},
  year    = {2024},
  volume  = {12},
  pages   = {157--173},
  doi     = {10.1162/tacl_a_00638}
}

@article{Salvatoreetal2025,
  author  = {Salvatore, Nikolaus and Wang, Hao and Zhang, Qiong},
  title   = {Lost in the Middle: An Emergent Property from Information Retrieval Demands in LLMs},
  journal = {arXiv},
  year    = {2025},
  doi     = {10.48550/arxiv.2510.10276}
}

@article{Cuconasuetal2025,
  author  = {Cuconasu, Florin and Filice, Simone and Horowitz, Guy and Maarek, Yoelle and Silvestri, Fabrizio},
  title   = {Do RAG Systems Really Suffer From Positional Bias?},
  journal = {arXiv},
  year    = {2025},
  doi     = {10.48550/arxiv.2505.15561}
}

@inproceedings{abnar-zuidema-2020-quantifying,
    title = "Quantifying Attention Flow in Transformers",
    author = "Abnar, Samira and Zuidema, Willem",
    booktitle = "Proceedings of the 58th Annual Meeting of the Association for Computational Linguistics",
    year = "2020",
    publisher = "Association for Computational Linguistics",
    pages = "4190--4197",
    doi = "10.18653/v1/2020.acl-main.385"
}
\clearpage

\appendix
\label{sec:appendix}
\renewcommand{\thefigure}{A\arabic{figure}}
\setcounter{figure}{0}
\renewcommand{\thesection}{\Alph{section}}
\setcounter{section}{0}
\section*{Appendix Overview}

\noindent The appendix is organized as follows:

\begin{itemize}
  \item[\textbf{\ref{app:related}}] \textbf{Related Work} — Extended discussion of positional bias in LLMs, mechanistic analyses of positional sensitivity, and layer-wise importance methods.
  \item[\textbf{\ref{app:datasets}}] \textbf{Dataset Details} — Descriptions of the eight texts across three genres and the scrambled control used in all experiments.
  \item[\textbf{\ref{app:wordtype}}] \textbf{Word-Type Analysis Details} — POS tagging categories and permutation test mechanics for the positional bias differential analysis.
  \item[\textbf{\ref{app:profiles}}] \textbf{Additional Positional Importance Profiles} — Layer-wise profiles for all texts not shown in the main paper (Shapes, Scrambled Pie Man, Wikipedia articles, Scientific abstracts).
  \item[\textbf{\ref{app:profiles:consistency}}] \textbf{Cross-Text Consistency} — Pairwise Pearson correlations of positional profiles across texts, confirming text-invariant structure.
  \item[\textbf{\ref{app:p50}}] \textbf{Robustness to Window Length ($P=50$)} — Replication of main analyses with a longer window, confirming findings are not an artifact of $P=10$.
  \item[\textbf{\ref{app:heatmaps}}] \textbf{Layer-Position Heatmaps} — Mean and variance conductance heatmaps across layers and positions for all texts and models.
\end{itemize}

\section{Related Work}
\label{app:related}

\subsection{Positional Bias in LLMs}

Language models often prioritize input position over semantic relevance. This is widely observed as the ``lost-in-the-middle'' effect: models struggle to access information located in the center of long contexts, heavily favoring the beginning (primacy) or end (recency) \cite{gao-etal-2024-insights, wang2025eliminating, hsieh-etal-2024-found, ravaut-etal-2024-context}. This bias degrades performance across standard NLP tasks, including question answering, summarization, and retrieval \cite{ko-etal-2020-look, shi2023large, Liuetal2024}.

However, the shape and severity of this positional bias vary significantly. First, the effect is model-specific; some architectures lean toward primacy, while others strongly favor recency \cite{zhang-etal-2024-bench, menschikov2025earlytokenbiasmodelspecificlanguagespecific}. Second, the bias shifts with context length. \citet{veseli2025positional} show that primacy peaks when relevant content occupies the first half of the window, whereas recency dominates longer inputs. Finally, the bias depends heavily on task demands. \citet{Salvatoreetal2025} show that the U-shaped performance curve is largely an artifact of information retrieval pressures during training. Similarly, \citet{Cuconasuetal2025} demonstrate that positional bias becomes marginal in practical retrieval-augmented generation (RAG) settings when distracting edge passages are equally penalized.

In short, prior work measures positional bias primarily through task-level outcomes. Because task difficulty and context length confound the model's intrinsic bias, these approaches cannot isolate how individual layers internally allocate positional importance.

\subsection{Mechanistic Analyses of Positional Structure}

Mechanistic studies probe transformers' internal computations to characterize how positional structure emerges in attention and representations. For instance, \citet{xiao2024efficient} and \citet{han2024lm} identify \textit{attention sinks}---specific positions, often the first or special tokens, that accumulate massive attention weight. Prioritizing empirical findings over theoretical labels, these studies demonstrate that such sinks drive the causal propagation of information across layers regardless of token semantics. 

Formalizing these internal dynamics, recent theoretical studies explicitly define this phenomenon as architectural positional bias. \citet{wu2025emergence} and \citet{herasimchyk2026residual} use graph-theoretic frameworks to prove how causal masking and residual connections inherently bias deeper layers toward earlier sequence positions. To track this bias across multiple layers, these studies rely on attention rollout \cite{abnar-zuidema-2020-quantifying}, a propagation-based approximation of attention distributions. Additionally, broader mechanistic reviews note that positional encodings (e.g., RoPE) enable sequence awareness but can induce superposition and interference in internal representations \cite{marks2025sparsefeaturecircuitsdiscovering}. 

However, attention-based methods (including rollout) only reveal internal routing; they do not measure a token's actual influence on the model's final prediction. \citet{pascual-etal-2021-telling} demonstrate this in BERT by comparing local attention to gradient attribution. They find a significant mismatch between the two caused by the extensive mixing of context across layers, although they observe that certain positional fingerprints persistently influence later layers. 

Overall, while mechanistic work confirms the architectural presence of positional structure, it stops short of systematically isolating input positions from lexical content to quantify their actual, layer-by-layer predictive contribution.

\subsection{Layer-wise Importance in LLMs}

A separate line of interpretability research analyzes layer-wise attribution and importance in transformers, though typically without isolating input position. For example, \citet{hou2023decodinglayersaliencylanguage} decode hidden layers back to token space to produce layer-specific saliency maps. \citet{ferrando-etal-2022-measuring} introduce ALTI, aggregating token-to-token interactions to measure layer-wise input attribution. More recently, \citet{kongmanee2025unravelingtokenpredictionrefinement} apply a logit-lens approach to trace how token predictions refine across depth, while \citet{ikeda2025layerwiseimportanceanalysisfeedforward} measure structural importance by redistributing feed-forward subnetworks during pretraining. 

While these methods successfully map how semantic information and specific token features propagate through the network, they inherently entangle a token's structural position with its lexical content. Overall, extracting a stable, position-dependent importance profile across layers---one that measures the actual predictive contribution of a relative position regardless of the words that occupy it---remains under-explored.

\section{Dataset Details}
\label{app:datasets}

We analyzed eight natural texts across three genres, plus one scrambled
control text.

\paragraph{Narrative Stories.}
We used two stories from the Narratives dataset
\citep{nastase2021narratives}: \emph{Pie Man} (957 words) and \emph{Shapes}
(910 words). This dataset contains naturalistic spoken narratives transcribed
to text. We used the original word order without additional preprocessing.

\paragraph{Encyclopedic Articles.}
Three articles randomly selected from the English Wikipedia Monthly dataset
\citep{wikipedia_kamali_2025}: \emph{Alabama}, \emph{Albedo},
\emph{Anarchism}, with 735, 310, and 703 words respectively after cleaning.

\paragraph{Scientific Abstracts.}
Three abstracts randomly selected from the SciDocs corpus
\citep{thakur2021beir}: \emph{The importance of drawing in the mechanical
design process}, \emph{Learning Agents for Uncertain Environments (Extended
Abstract)}, and \emph{The Intersecting Roles of Consumer and Producer}, with
794, 739, and 793 words respectively.

\paragraph{Scrambled Control.}
We created a scrambled version of the \emph{Pie Man} story by uniformly
shuffling its word sequence. This control tests whether positional patterns
persist independent of meaningful content.


\section{Word-Type Analysis Details}
\label{app:wordtype}

\subsection{Part-of-Speech Categorization}
To group tokens by syntactic role, we processed all input texts using the spaCy \texttt{en\_core\_web\_sm} pipeline. Tokens were categorized into two broad classes based on their universal Part-of-Speech (POS) tags:
\begin{itemize}
    \item \textbf{Content Words:} Nouns (\texttt{NOUN}, \texttt{PROPN}), Verbs (\texttt{VERB}), Adjectives (\texttt{ADJ}), Adverbs (\texttt{ADV}), Interjections (\texttt{INTJ}), and Numerals (\texttt{NUM}).
    \item \textbf{Function Words:} Auxiliaries (\texttt{AUX}), Adpositions (\texttt{ADP}), Pronouns (\texttt{PRON}), Determiners (\texttt{DET}), Conjunctions (\texttt{CCONJ}, \texttt{SCONJ}), and Particles (\texttt{PART}).
\end{itemize}
Punctuation (\texttt{PUNCT}) and unrecognized tokens (\texttt{X}) were excluded. Prior to analysis, all valid tokens were pooled across the input texts. We excluded the scrambled control text from this pool because it is a lexically identical permutation of the \textit{Pie Man} story; including it would redundantly double-count those specific words.

\subsection{Permutation Test Mechanics}
Because our methodology requires aggregating token conductance to form a single positional importance profile per word type before extracting the primacy and recency fractions, token-level variance is collapsed. This renders standard parametric tests (e.g., $t$-tests) invalid. 

To assess the statistical significance of the difference between word types ($\Delta = \text{content} - \text{function}$), we utilized a non-parametric layer-wise permutation test. For each layer of each model, we performed 1,000 iterations. In each iteration, the word-type labels were randomly shuffled across the pooled valid tokens, preserving the true marginal frequencies of the dataset ($N_{\text{content}}$ and $N_{\text{function}}$). 

For each randomized permutation, the aggregate positional importance profiles and resulting primacy/recency fractions were recomputed to construct a null distribution of $\Delta$ values. The true observed $\Delta$ was then compared against this null distribution to compute a two-sided $p$-value. To account for multiple comparisons across layers, these $p$-values were adjusted using the Benjamini-Hochberg False Discovery Rate (FDR) correction. 

Finally, to ensure our reported findings generalize across transformer families rather than reflecting model-specific artifacts, we applied a cross-model consensus threshold. A layer depth is only reported as statistically significant if the metric passed the FDR-corrected $p < 0.05$ threshold and exhibited the same directionality ($\Delta > 0$ or $\Delta < 0$) independently across all four evaluated models (GPT-2, GPT-Neo, Phi-2, Llama-3).
\clearpage
\section{Additional Positional Importance Profiles}
\label{app:profiles}

\begin{figure}[h]
  \centering
  \includegraphics[width=1\linewidth]{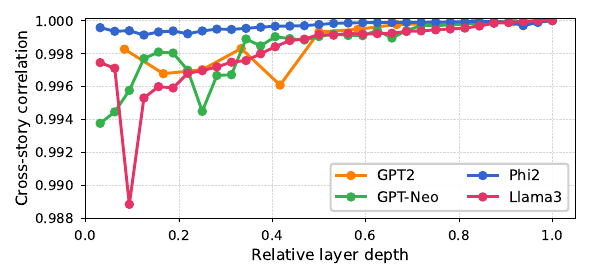}
  \vspace{-3mm}
  \caption{Cross-text consistency of positional importance profiles. Mean
    pairwise Pearson correlation between profiles across inputs for each
    layer. Correlations exceed $r > 0.99$ across all models, indicating
    text-invariant positional profiles.}
  \label{fig:story_invariance_appendix}
  \vspace{-5mm}
\end{figure}

This section reports the layer-wise positional importance profiles for all
texts not shown in the main paper. 
(Figure~\ref{fig:story_invariance_appendix}) holds for every text examined.
\subsection{Narrative Stories}
\label{app:profiles:narratives}

\begin{figure*}[b]
  \centering
  \includegraphics[width=0.8\linewidth]{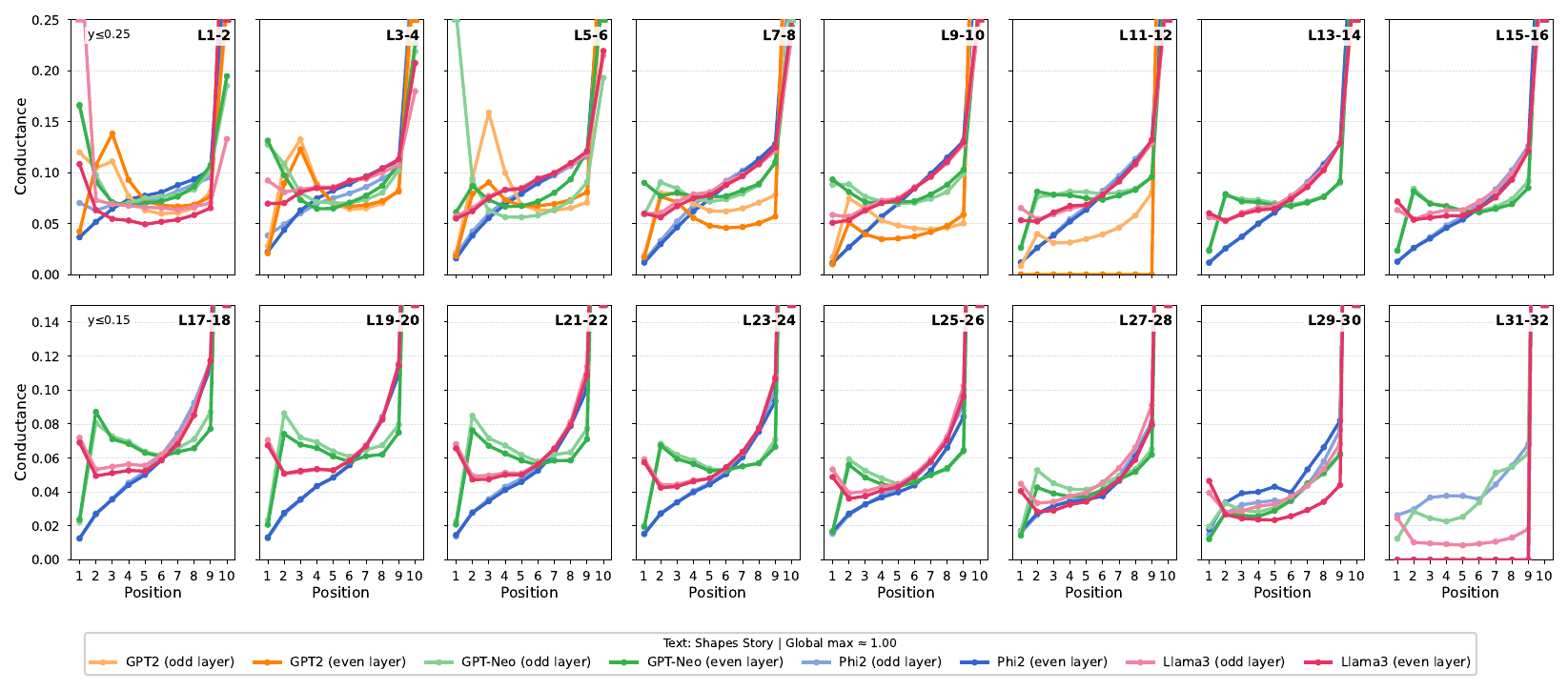}
  \caption{\textbf{Shapes:} Layer-wise positional importance profiles.
    Conductance scores averaged over words as a function of position.
    Patterns are consistent with the main text (Figure~2).}
  \label{fig:layer_profiles_shapes}
\end{figure*}

\FloatBarrier

\subsection{Scrambled Control}
\label{app:profiles:scrambled}

\begin{figure*}[b]
  \centering
  \includegraphics[width=0.8\linewidth]{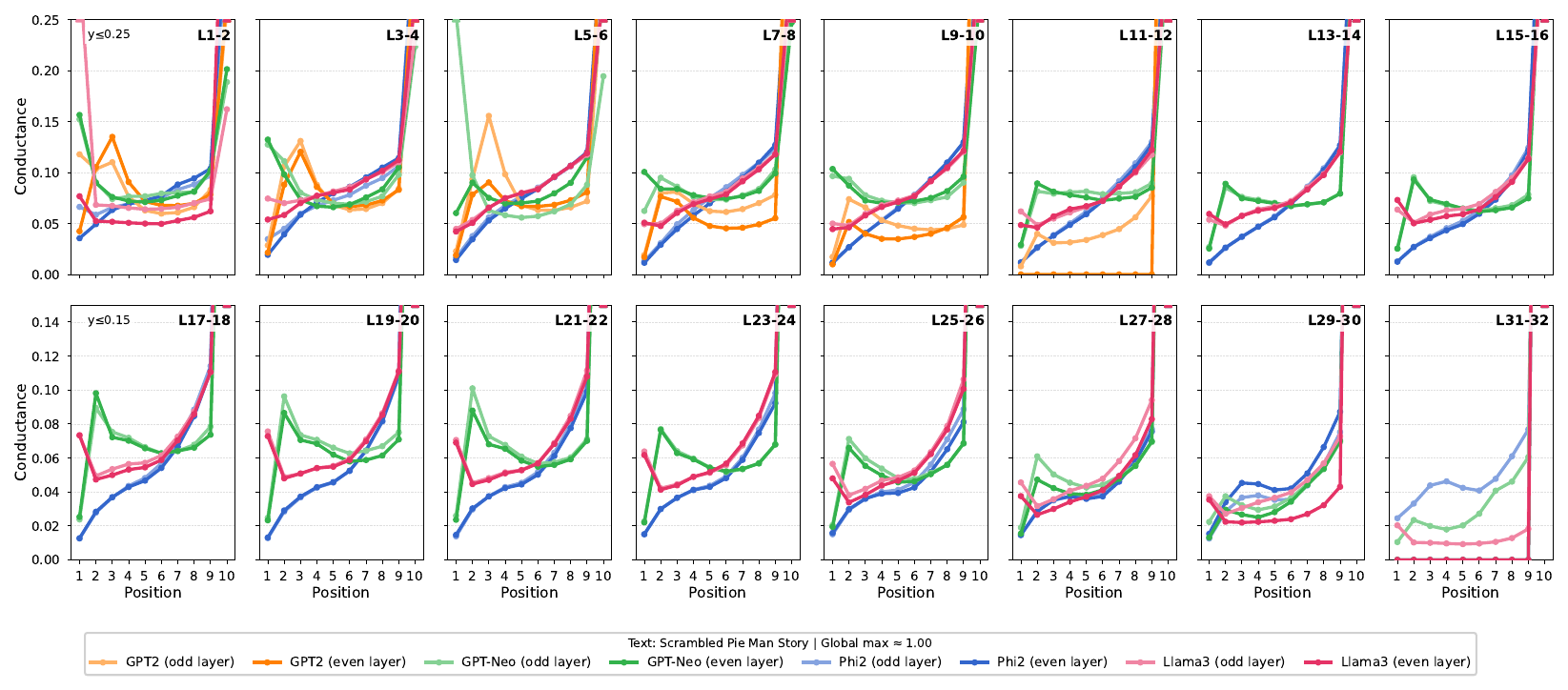}
  \caption{\textbf{Scrambled Pie Man:} Layer-wise positional importance
    profiles. Despite scrambling, similar recency and primacy peaks are
    observed, confirming architectural origin.}
  \label{fig:layer_profiles_scrambled}
\end{figure*}

\clearpage
\subsection{Wikipedia Articles}
\label{app:profiles:wikipedia}
\vspace{0.5cm} 
\begin{figure*}[b]
  \centering
  
  \includegraphics[width=0.6\linewidth]{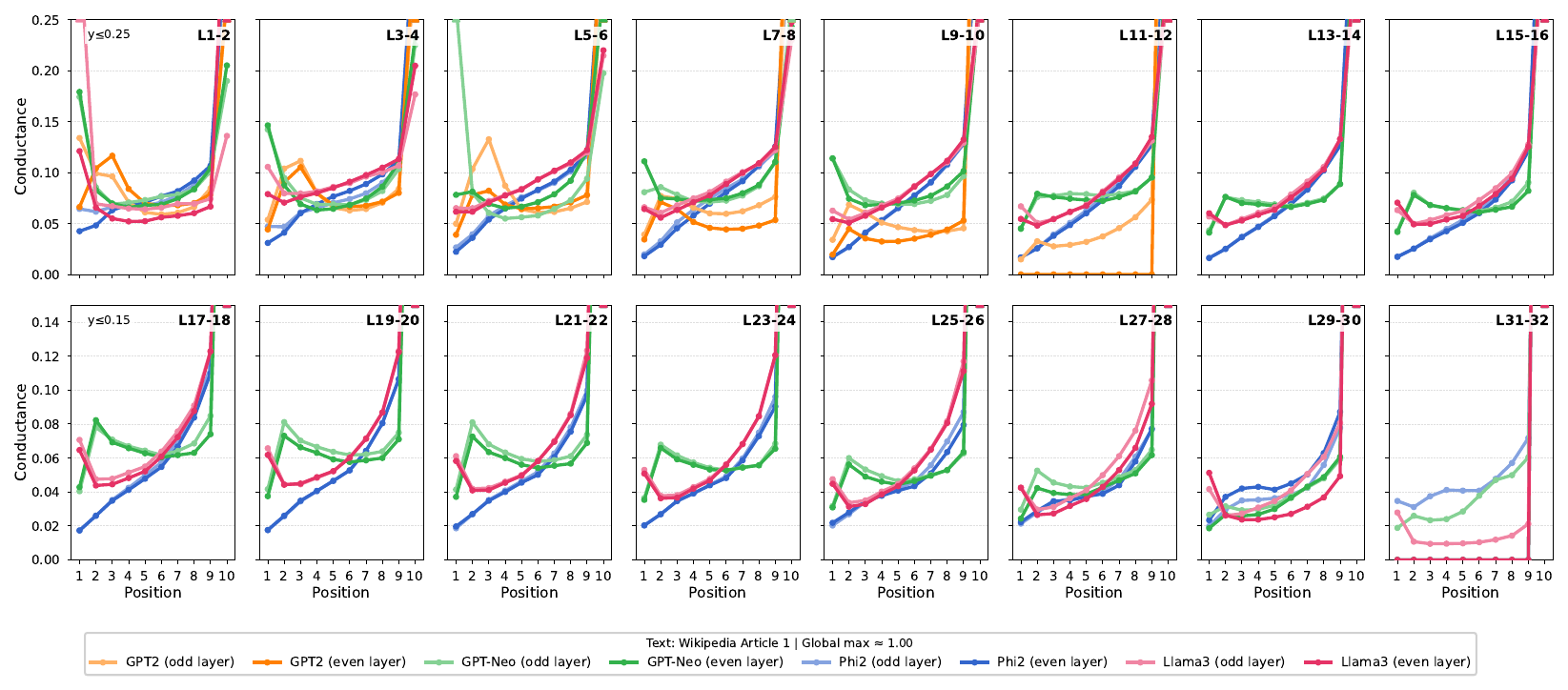}
  \caption{\textbf{Wikipedia Article 1:} Layer-wise positional importance profiles.}
  \label{fig:layer_profiles_wiki1}
  
  \vspace{0.5cm} 
  
  \includegraphics[width=0.6\linewidth]{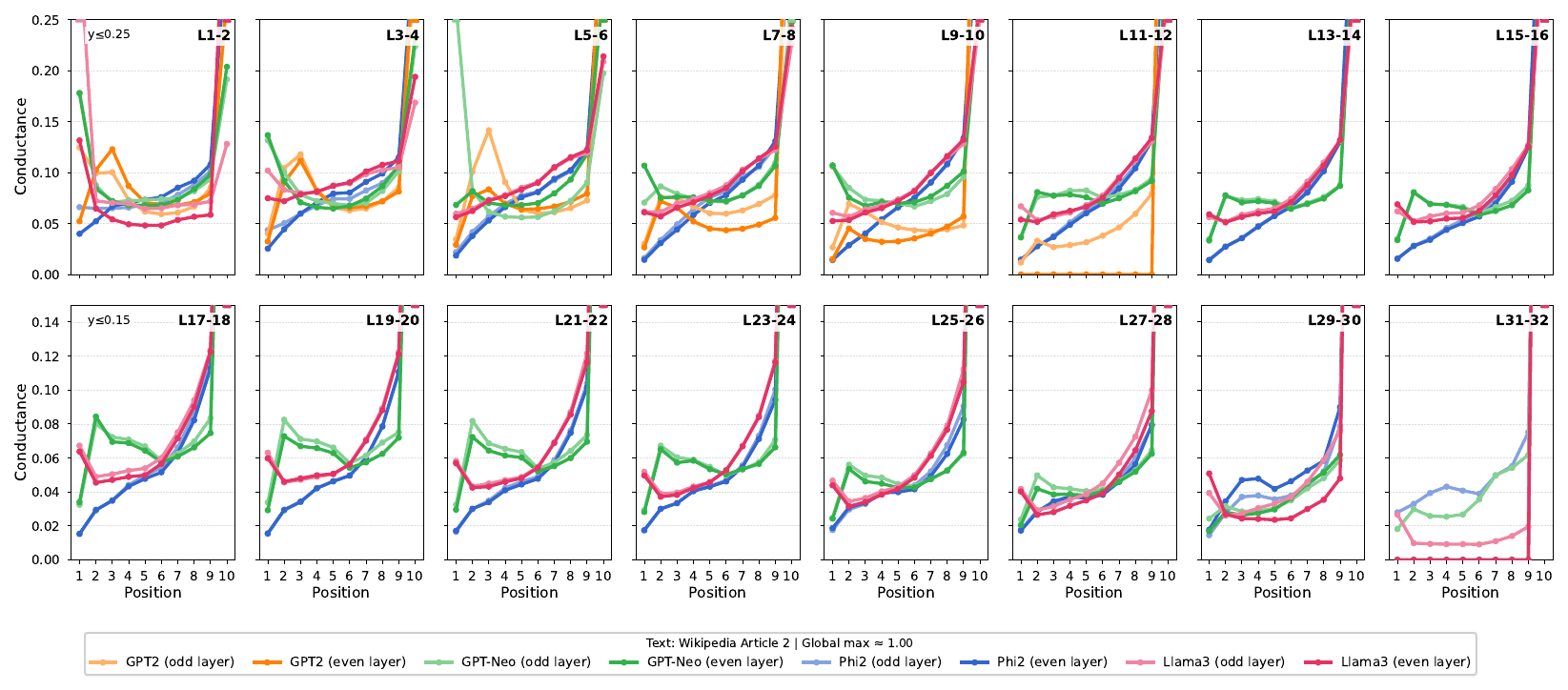}
  \caption{\textbf{Wikipedia Article 2:} Layer-wise positional importance profiles.}
  \label{fig:layer_profiles_wiki2}
  
  \vspace{0.5cm}
  
  \includegraphics[width=0.6\linewidth]{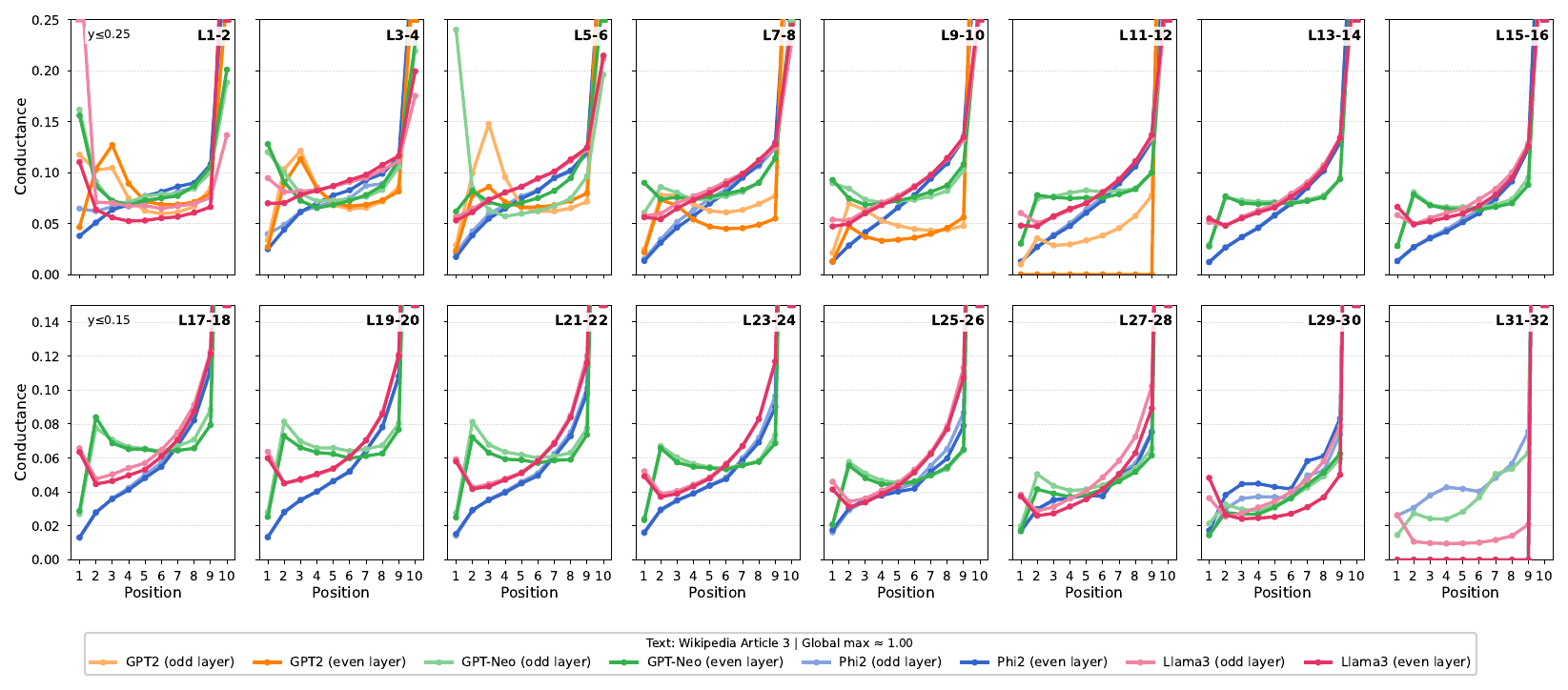}
  \caption{\textbf{Wikipedia Article 3:} Layer-wise positional importance profiles.}
  \label{fig:layer_profiles_wiki3}
\end{figure*}
\FloatBarrier
\clearpage
\subsection{Scientific Abstracts}
\label{app:profiles:scientific}
 \vspace{2cm}
\begin{figure*}[b]
  \centering
  \includegraphics[width=0.6\linewidth]{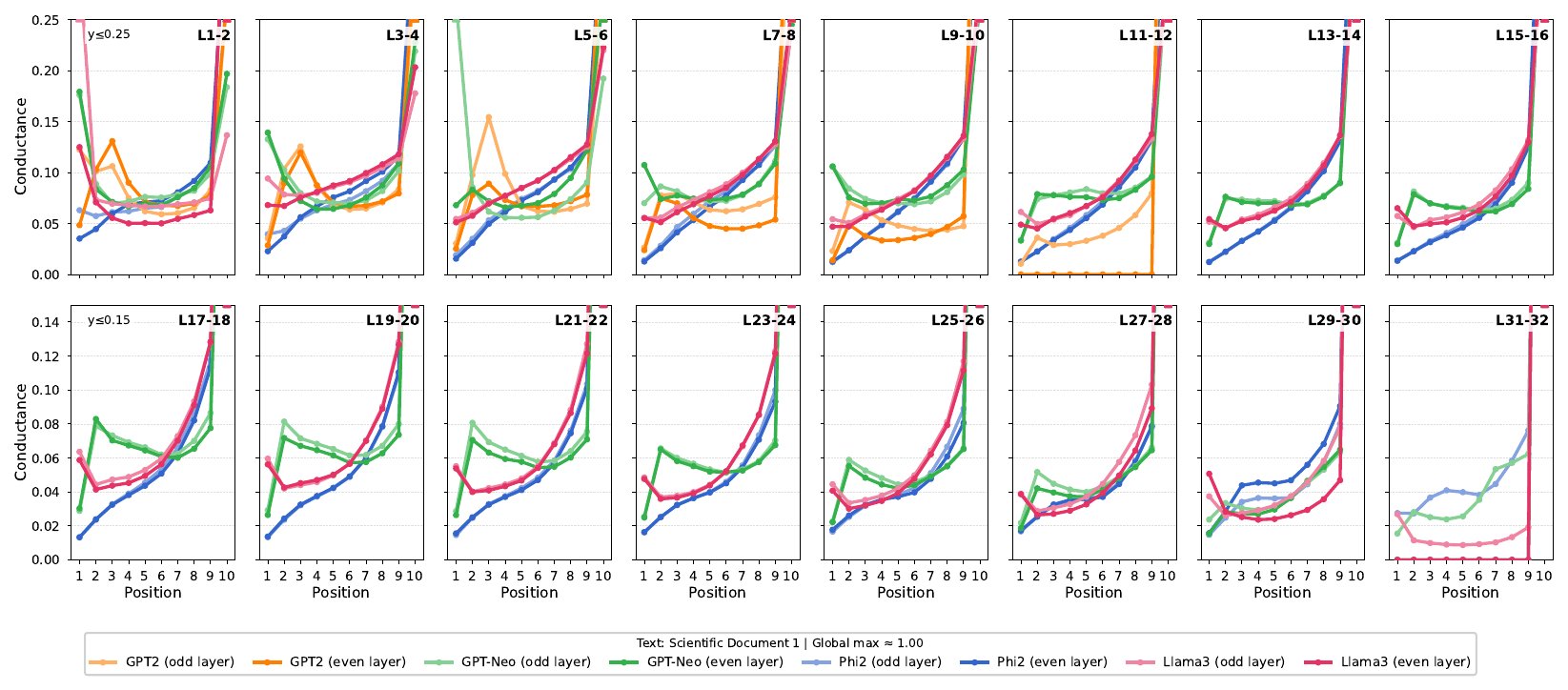}
  \caption{\textbf{Scientific Document 1:} Layer-wise positional importance profiles.}
  \label{fig:layer_profiles_scidoc1}

 \vspace{0.5cm}

  \centering
  \includegraphics[width=0.6\linewidth]{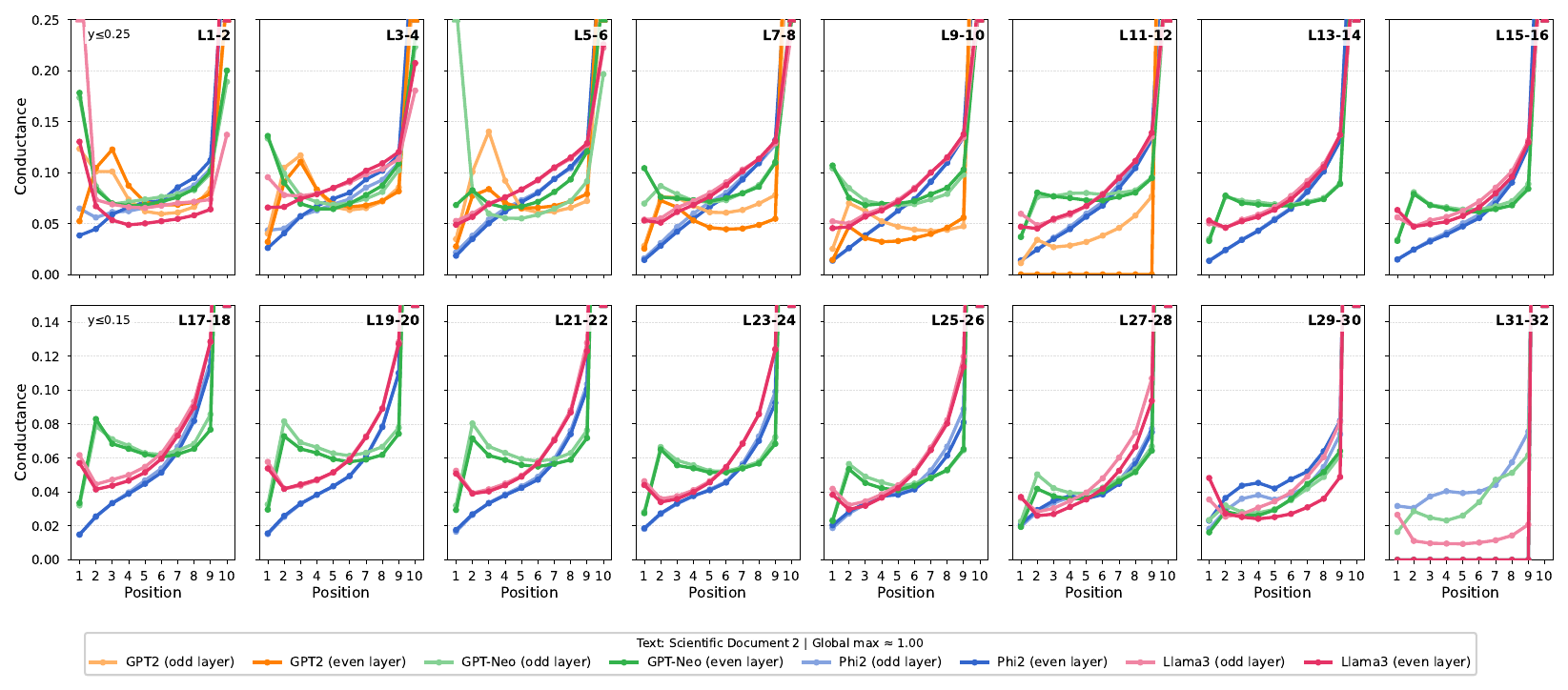}
  \caption{\textbf{Scientific Document 2:} Layer-wise positional importance profiles.}
  \label{fig:layer_profiles_scidoc2}
 \vspace{0.5cm}

  \centering
  \includegraphics[width=0.6\linewidth]{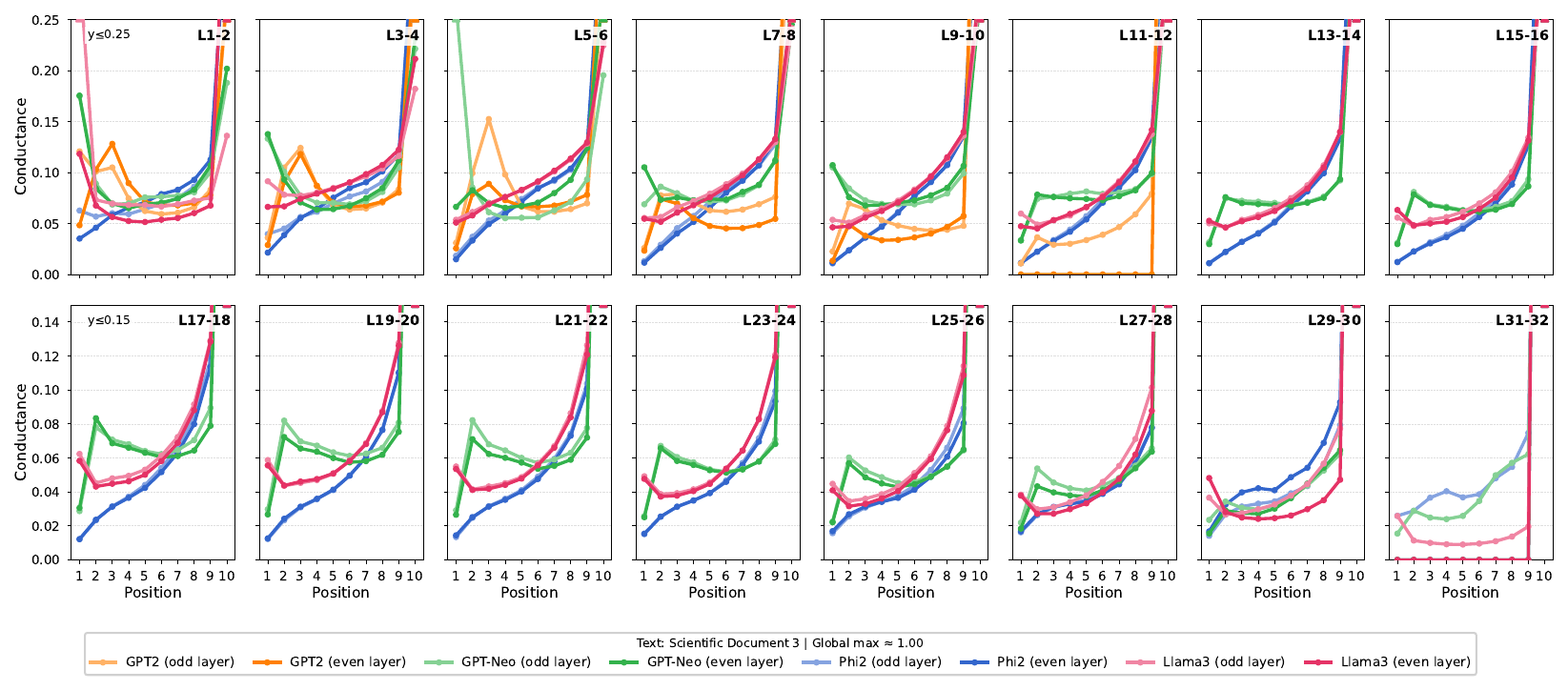}
  \caption{\textbf{Scientific Document 3:} Layer-wise positional importance profiles.}
  \label{fig:layer_profiles_scidoc3}
\end{figure*}

\FloatBarrier
\clearpage
\section{Cross-Text Consistency}
\label{app:profiles:consistency}

\section{Robustness to Window Length: Analysis with $P=50$}
\label{app:p50}

To verify that our findings are robust to window length, we repeated core
analyses with $P=50$ on two representative texts: \emph{Pie Man} and
\emph{Shapes}. The key patterns replicate with the longer window, demonstrating that this positional bias reflects robust model-internal behavior independent of window size.

\subsection{Positional Importance Profiles}
\label{app:p50:profiles}
\vspace{-4mm}
\begin{figure*}[b]
  \centering
  \includegraphics[width=1\linewidth]{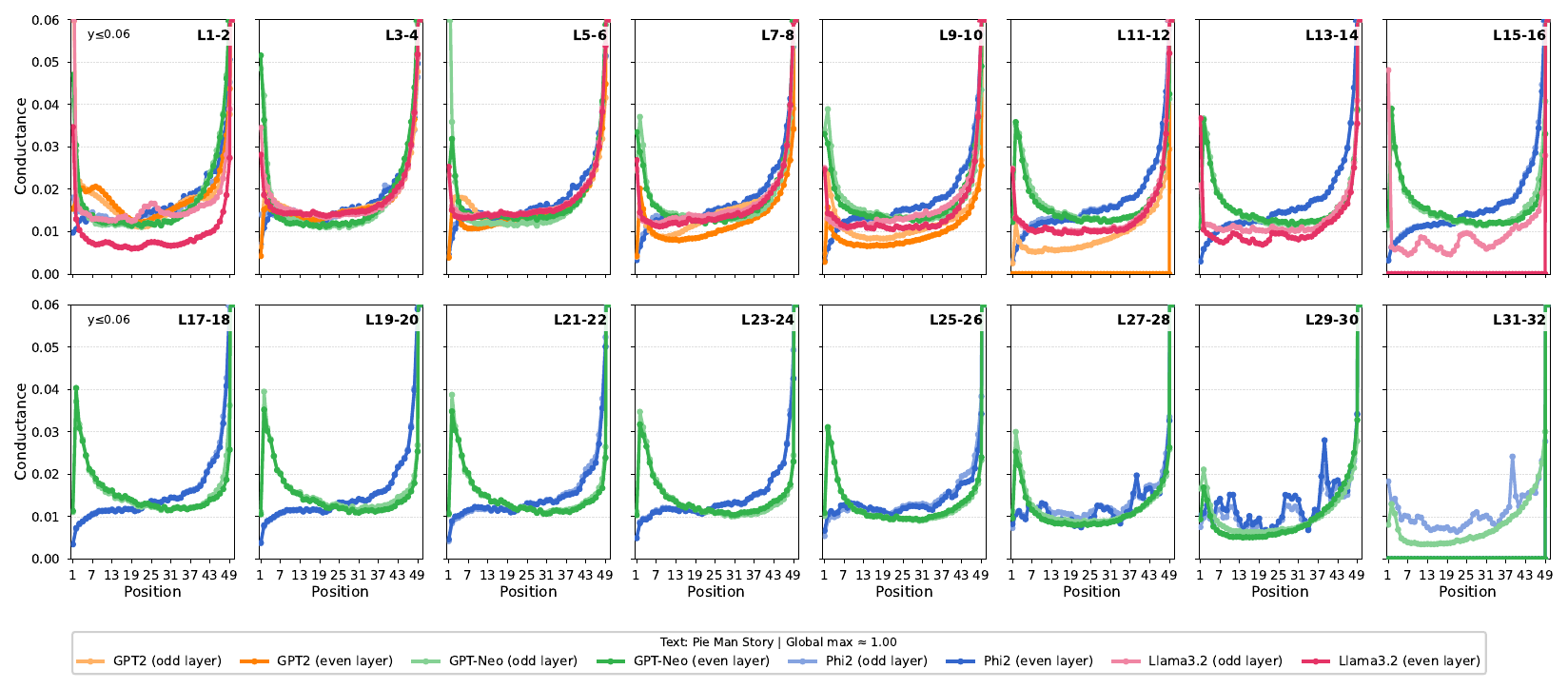}
  \caption{\textbf{Pie Man ($P=50$):} Layer-wise positional importance
    profiles. Patterns replicate those observed with $P=10$ (main text,
    Figure~2).}
  \label{fig:layer_profiles_pieman_50}
\end{figure*}
\begin{figure*}[h]
  \centering
  \includegraphics[width=1\linewidth]{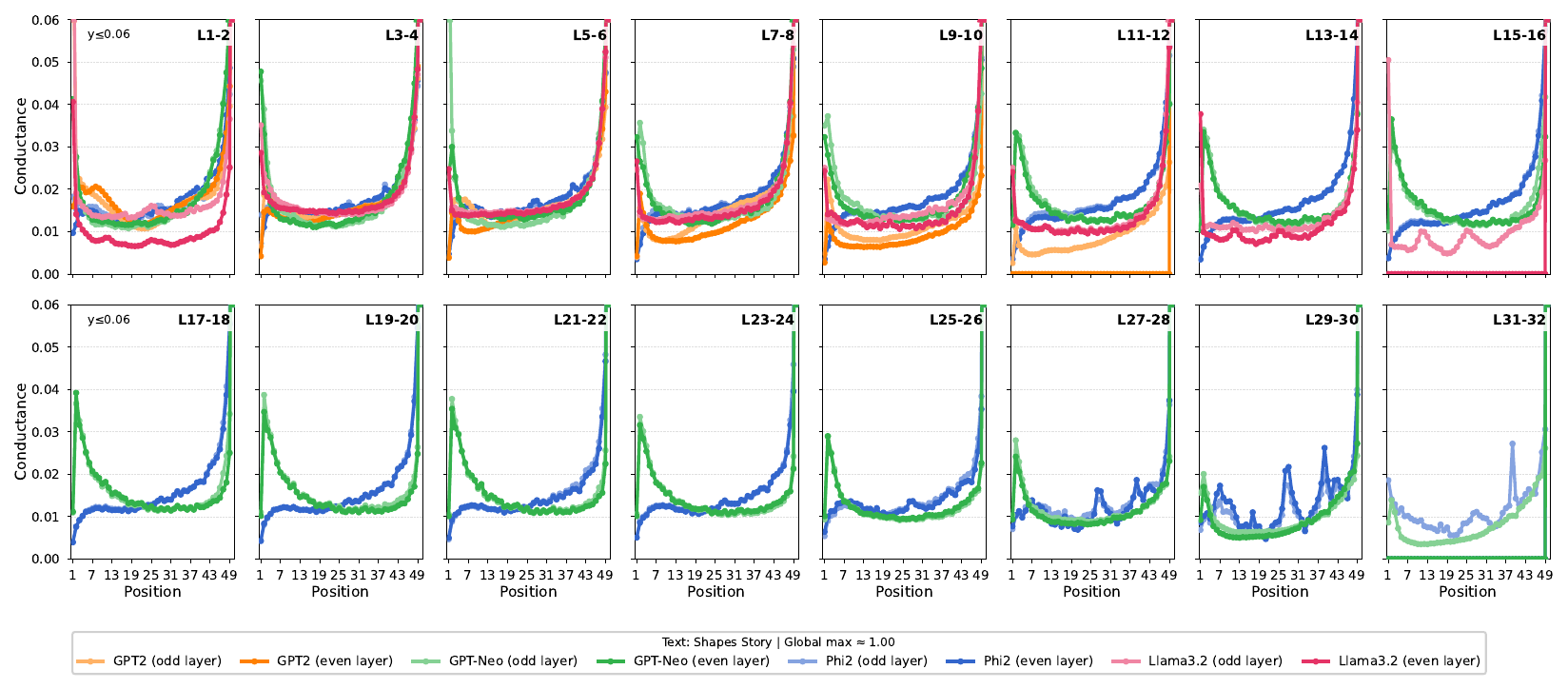}
  \caption{\textbf{Shapes ($P=50$):} Layer-wise positional importance
    profiles. Patterns consistent with $P=10$ analysis.}
  \label{fig:layer_profiles_shapes_50}

  \centering

  \includegraphics[width=1\linewidth]{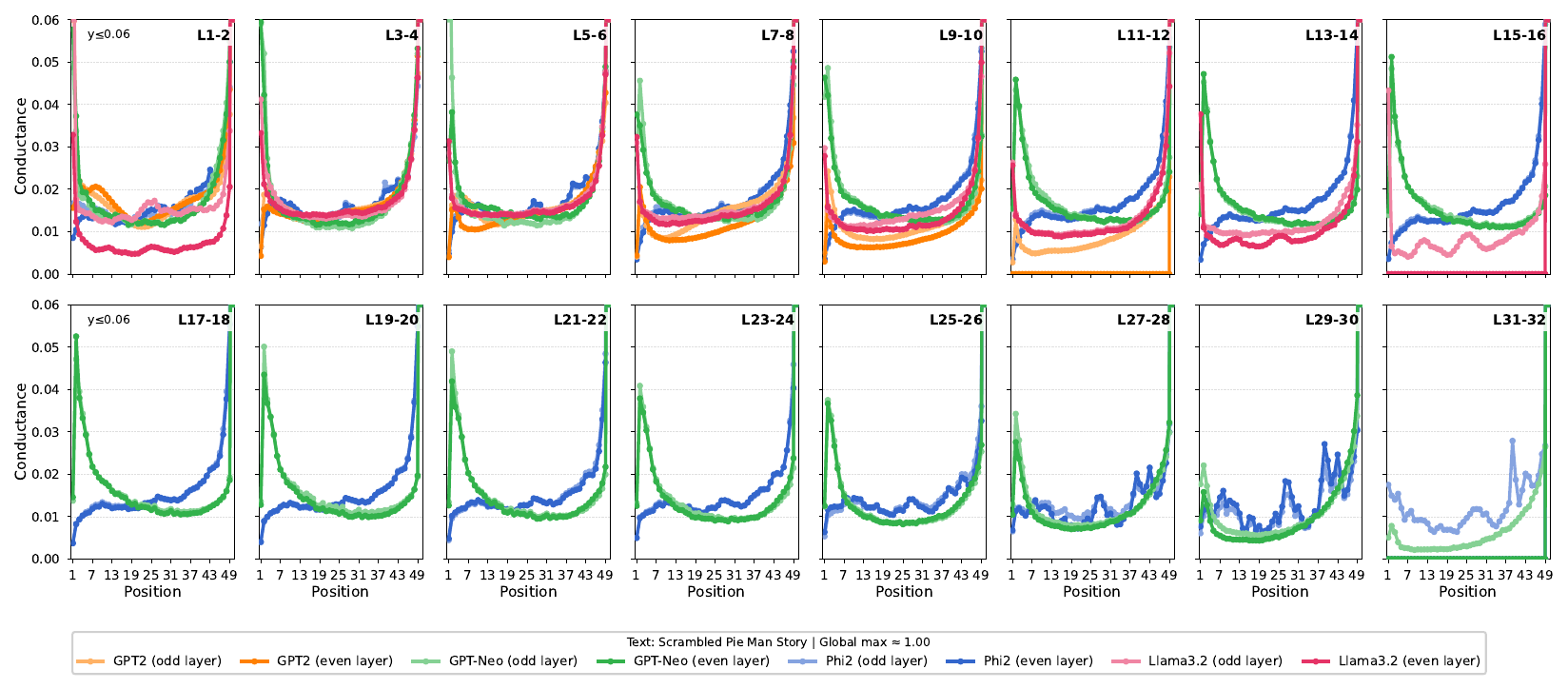}
  \caption{\textbf{Scrambled Pie Man ($P=50$):} Positional patterns persist
    under scrambling, as with $P=10$.}
  \label{fig:layer_profiles_scrambled_50}
    \vspace{-5mm}
\end{figure*}

\FloatBarrier
\subsection{Cross-Text Consistency}
\label{app:p50:consistency}
\begin{figure}[H]
  \centering
  \includegraphics[width=0.7\linewidth]{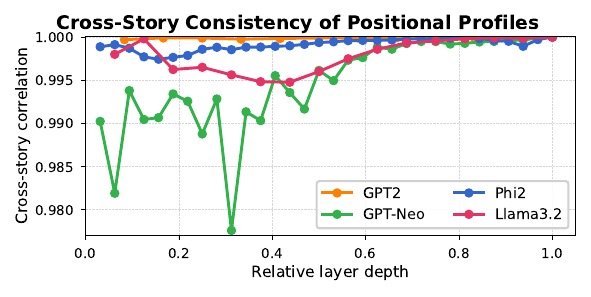}
  \vspace{-3mm}
  \caption{\textbf{Cross-text consistency ($P=50$).} Mean pairwise Pearson
    correlation between positional profiles across texts for each layer.
    Correlations exceed $r > 0.99$ across all models, replicating
    text-invariance with longer windows.}
  \label{fig:story_invariance_50}
  \vspace{-5mm}
\end{figure}


\subsection{Primacy and Recency Evolution}
\label{app:p50:evolution}

\begin{figure}[H]
  \centering
  \includegraphics[width=0.7\linewidth]{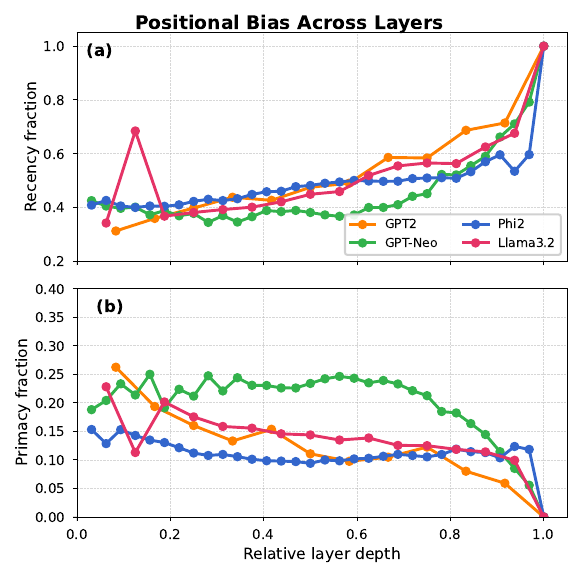}
  \vspace{-3mm}
  \caption{\textbf{Primacy and recency evolution ($P=50$).} (a) Recency
    fraction; (b) Primacy fraction. Patterns averaged across texts. Recency
    increases monotonically with depth; primacy remains weak and diminishing,
    replicating $P=10$ findings.}
  \label{fig:primacy_recency_50}
  \vspace{-5mm}
\end{figure}

\subsection{Positional Bias Across Word Types}
\begin{figure}[h]
    \centering
    \includegraphics[width=1\linewidth]{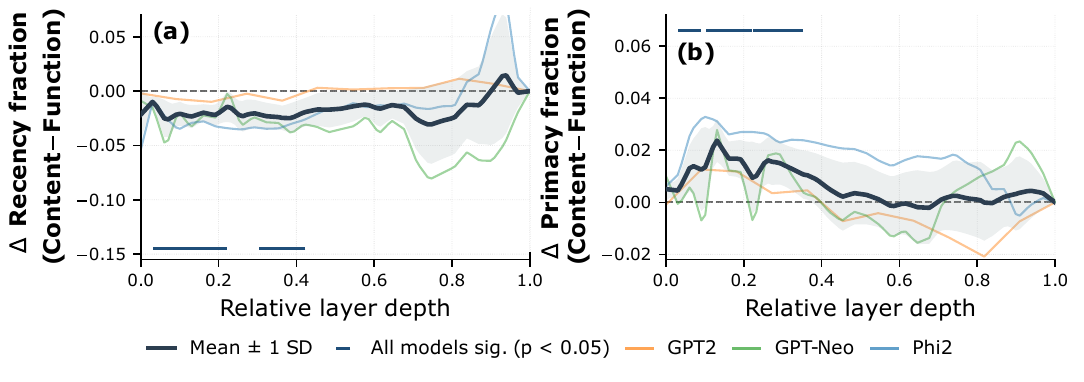}


\caption{Per-layer difference in PrimFrac and RecFrac between content and function words ($\Delta$ = content $-$ function) ($P=50$). Words are pooled across all texts (excluding scrambled). Bold line: cross-model mean $\pm$1 SD. Horizontal bar: all three evaluated models (GPT-2, GPT-Neo, Phi-2) are significant in the same direction (permutation test, FDR-corrected $p < 0.05$). Note: Llama-3.2 is omitted from this specific extended-context replication due to file availability constraints at submission; however, the highly consistent transition observed across the other architectures demonstrates this pattern is robust.}

     \label{fig:app_wordtype_delta}
\end{figure}

\subsection{Summary}

The key findings from $P=10$ replicate with $P=50$: text-invariant profiles ($r > 0.99$), monotonic recency increase with depth, weak and diminishing primacy, and significant word-type sensitivity ($\Delta$ fractions) isolated to the early layers. This confirms that the observed positional structure and early lexical processing reflect stable model-internal routing behavior, rather than an artifact of the chosen window length.

\section{Layer-Position Heatmaps}
\label{app:heatmaps}

The following heatmaps provide an alternative view of positional importance,
showing mean and variance of conductance across layers (rows) and positions
(columns) for each model and text. They complement the profile plots above and
may be useful for visually identifying model-specific structure.
\clearpage

\subsection{Mean Conductance}
\label{app:heatmaps:mean}

\begin{figure*}[b]
  \centering
  \includegraphics[width=1\linewidth]{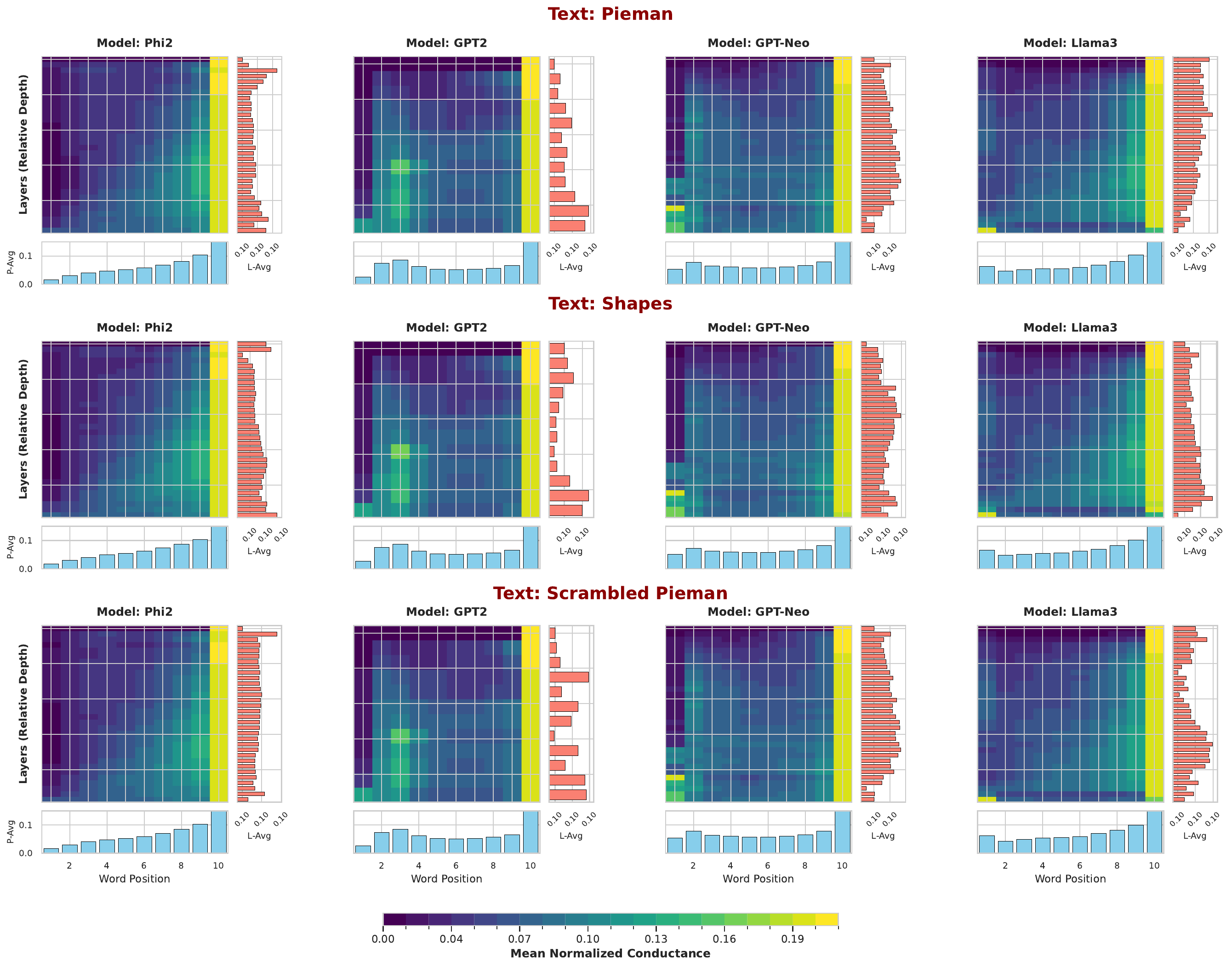}
  \caption{Mean conductance heatmaps — Narrative texts.
    Rows: layers; columns: positions.}
  \label{fig:heatmap_mean1}
\end{figure*}

\begin{figure*}[h]
  \centering
  \includegraphics[width=\linewidth]{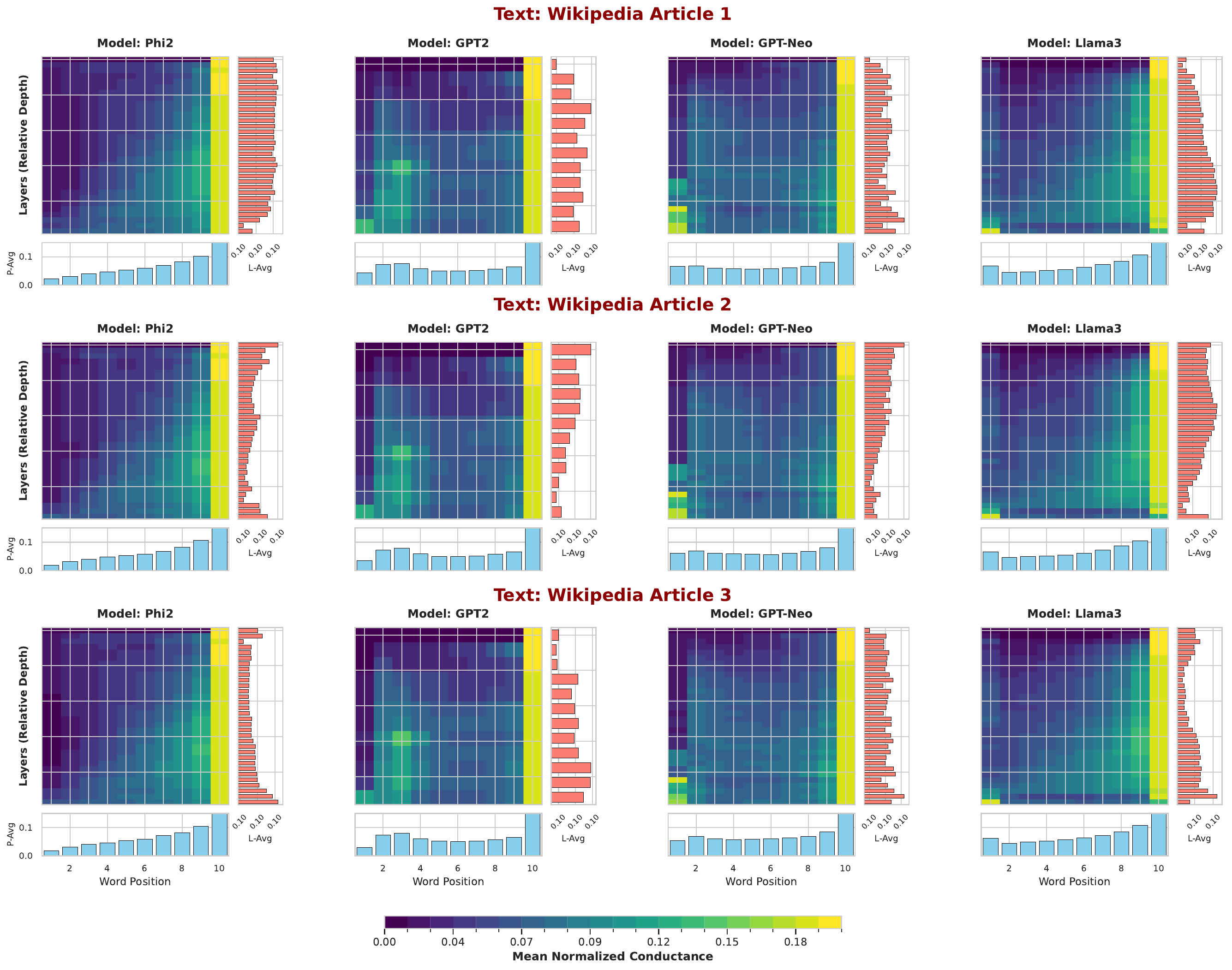}
  \caption{Mean conductance heatmaps — Wikipedia articles.}
  \label{fig:heatmap_mean2}
\end{figure*}

\begin{figure*}[h]
  \centering
  \includegraphics[width=\linewidth]{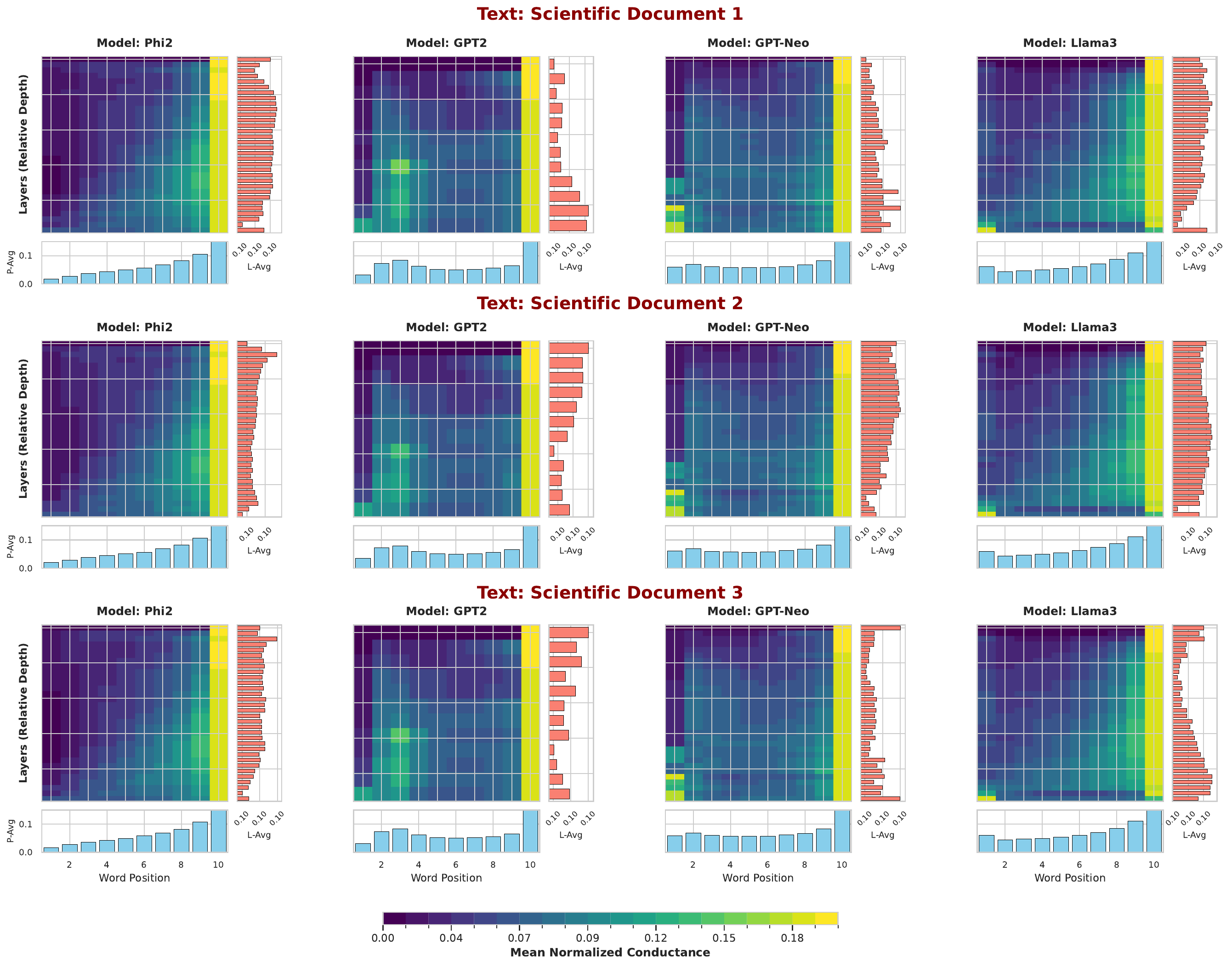}
  \caption{Mean conductance heatmaps — Scientific abstracts.}
  \label{fig:heatmap_mean3}
\end{figure*}

\FloatBarrier
\clearpage

\subsection{Variance in Conductance}
\label{app:heatmaps:variance}

\begin{figure*}[b]
  \centering
  \includegraphics[width=\linewidth]{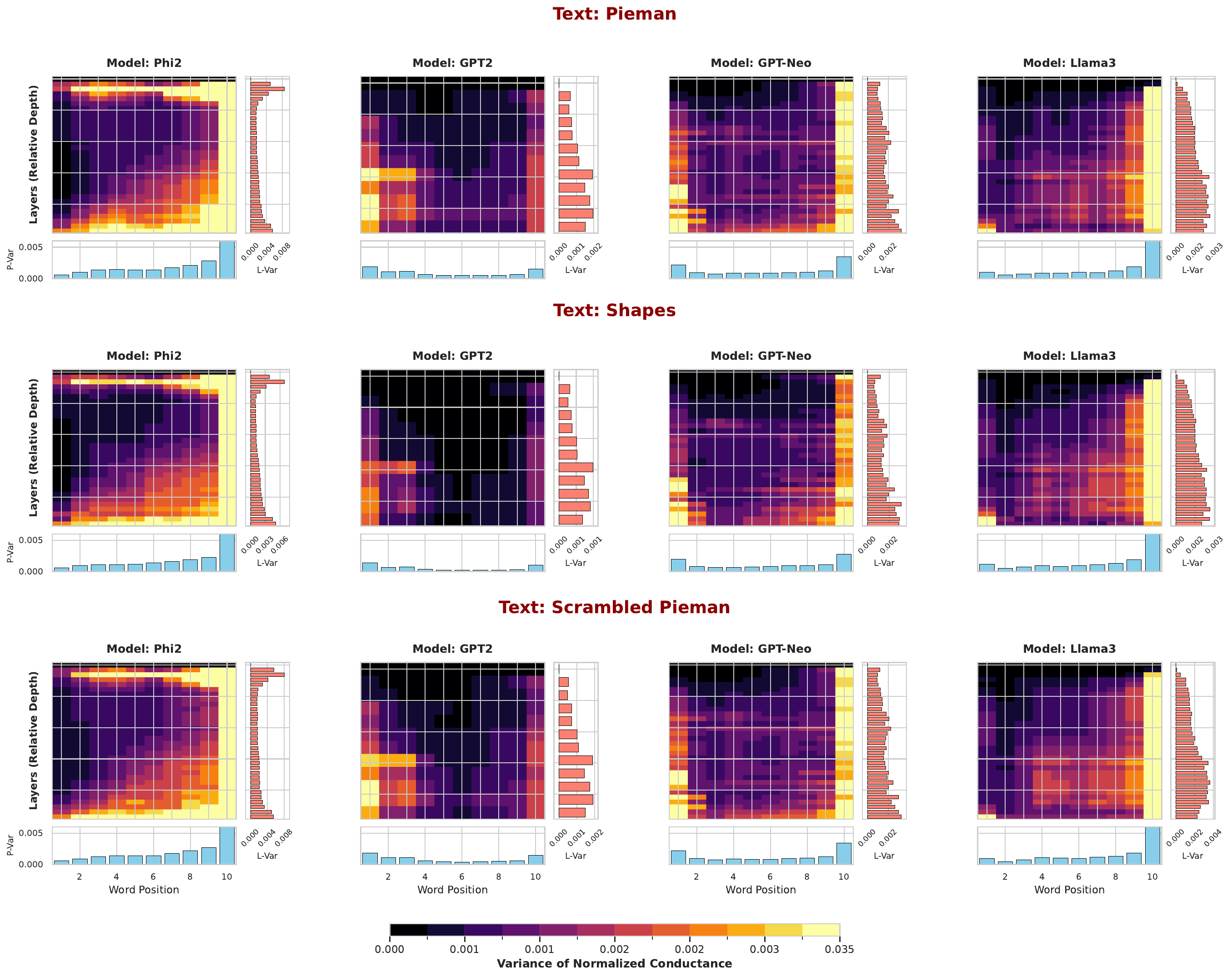}
  \caption{Variance in conductance — Narrative texts.
    Shows positional stability of layer contributions.}
  \label{fig:heatmap_variance1}
\end{figure*}

\begin{figure*}[h]
  \centering
  \includegraphics[width=\linewidth]{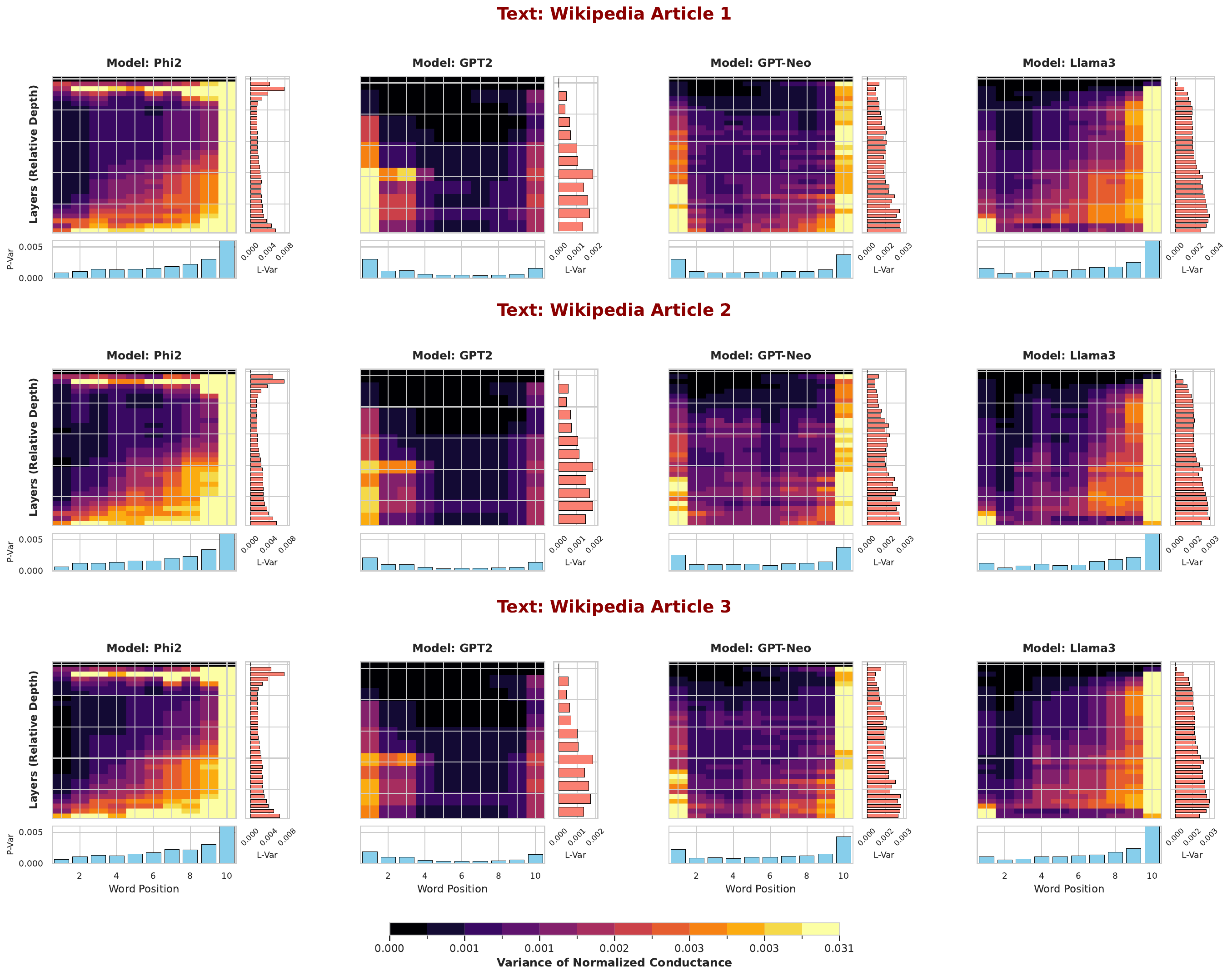}
  \caption{Variance in conductance — Wikipedia articles.}
  \label{fig:heatmap_variance2}
\end{figure*}

\begin{figure*}[h]
  \centering
  \includegraphics[width=\linewidth]{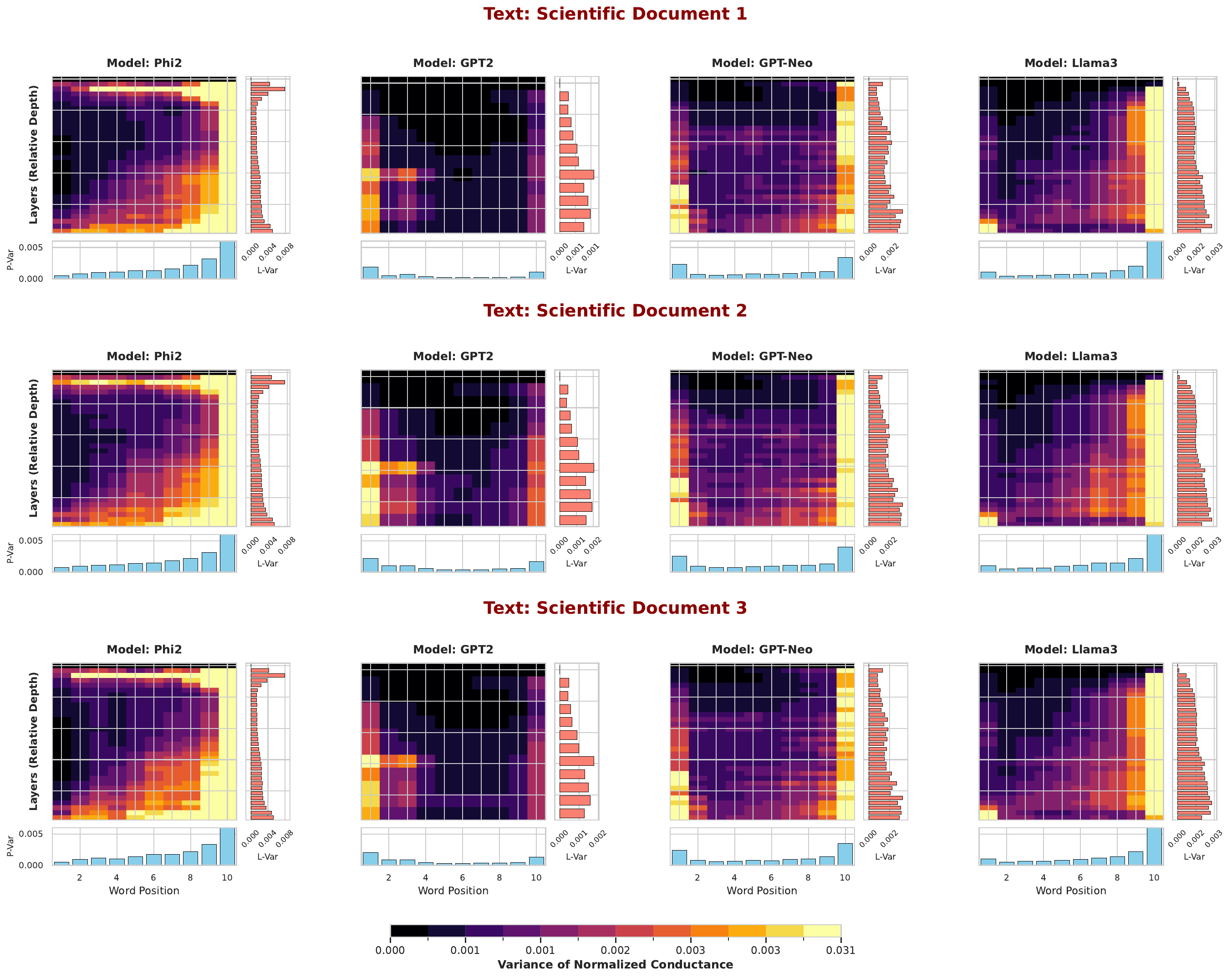}
  \caption{Variance in conductance — Scientific abstracts.}
  \label{fig:heatmap_variance3}
\end{figure*}

\FloatBarrier
\clearpage

\end{document}